\documentclass[letterpaper, 10 pt, conference]{ieeeconf}

\IEEEoverridecommandlockouts                              %

\usepackage{amsmath,amsfonts}
\usepackage{algorithmic}
\usepackage{array}
\usepackage[caption=false,font=normalsize,labelfont=sf,textfont=sf]{subfig}
\usepackage{textcomp}
\usepackage{stfloats}
\usepackage{url}
\usepackage{verbatim}
\usepackage{graphicx}
\usepackage{hyperref}
\def\BibTeX{{\rm B\kern-.05em{\sc i\kern-.025em b}\kern-.08em
    T\kern-.1667em\lower.7ex\hbox{E}\kern-.125emX}}
\usepackage{balance}

\usepackage[font=small]{caption}
\usepackage{wrapfig}
\usepackage{xcolor}
\usepackage{csquotes}

\usepackage{moreverb,url}
\usepackage{amsfonts,amssymb,amsmath,bm, mathabx}
\usepackage[ruled,vlined]{algorithm2e}
\usepackage{multirow}
\usepackage{tabularx}
\usepackage{tabularray}
\UseTblrLibrary{siunitx}

\usepackage{cleveref}
\Crefformat{figure}{#2Fig.~#1#3}
\Crefformat{appendix}{#2App.~#1#3}

\usepackage{cleveref}

\usepackage[]{glossaries}
\newacronym{ipa}{IPA}{Invariant Point Attention}

\newacronym{dgm}{DGM}{Deep Generative Models}

\newacronym{gail}{GAIL}{Generative Adversarial Imitation Learning}
\newacronym{sqil}{SQIL}{Soft-Q Imitation Learning}

\newacronym{iflow}{iFlows}{ImitationFlows}
\newacronym{cep}{CEP}{Composable Energy Policies}
\newacronym{se3dif}{SE(3)-DiF}{SE(3)-DiffusionFields} %

\newacronym{nlp}{NLP}{Natural Language Processing}
\newacronym{cv}{CV}{Computer Vision}

\newacronym{gp}{GP}{Gaussian Process}
\newacronym{gan}{GAN}{Generative Adversarial Networks}
\newacronym{vae}{VAE}{Variational Autoencoders}
\newacronym{cvae}{cVAE}{Conditional Variational Autoencoders}
\newacronym{ebm}{EBM}{Energy Based Models}
\newacronym{sbm}{SBM}{Score based Models}
\newacronym{nf}{NFlow}{Normalizing Flows}

\newacronym{dm}{DM}{Diffusion Models}
\newacronym{fm}{FM}{Flow Matching}
\newacronym{cfm}{CFM}{Conditional Flow Matching}

\newacronym{ot}{OT}{Optimal Transport}

\newacronym{ddpm}{DDPM}{Denoising Diffusion Probabilistic Models}
\newacronym{ncsn}{NCSN}{Noise Conditioned Score Network}
\newacronym{smld}{SMLD}{Score Matching with Langevin Dynamics}

\newacronym{map}{MAP}{Maximum a Posteriori}

\newacronym{mdn}{MDN}{Mixture Density Networks}
\newacronym{gmm}{GMM}{Gaussian Mixture Models}

\newacronym{nerf}{NeRF}{Neural Radiance Fields}

\newacronym{bc}{BC}{Behavioural Cloning}
\newacronym{il}{IL}{Imitation Learning}
\newacronym{irl}{IRL}{Inverse Reinforcement Learning}
\newacronym{ioc}{IOC}{Inverse Optimal Control}
\newacronym{lfd}{LfD}{Learning from Demonstration}
\newacronym{em}{EM}{Expectation Maximization}
\newacronym{promp}{ProMP}{Probabilistic Movement Primitives}
\newacronym{dmp}{DMP}{Dynamic Movement Primitives}
\newacronym{seds}{SEDS}{Stable Estimator of Dynamical Systems}
\newacronym{gmr}{GMR}{Gaussian Mixture Regressor}
\newacronym{gpr}{GPR}{Gaussian Process Regressor}
\newacronym{lwr}{LWR}{Locally Weighted Regressor}
\newacronym{kmp}{KMP}{Kernelized Movement Primitives}
\newacronym{clf}{CLF}{Control Lyapunov Function}
\newacronym{wsaqf}{WSAQF}{Weighted Sum of Asymmetric Quadratic Function)}
\newacronym{nilc}{NILC}{Neurally Imprinted Lyapunov
Candidate}
\newacronym{clfdm}{CLF-DM}{Control Lyapunov Function-based Dynamic Movements}
\newacronym{cnmp}{CNMP}{Conditional Neural Movement Primitives}
\newacronym{tpgmm}{TP-GMM}{Task Parameterized GMM}

\newacronym{mp}{MP}{Movement Primitive}
\newacronym{mpflows}{MPFlows}{Movement Primitive Flows}

\newacronym{gcl}{GCL}{Guided Cost Learning}

\newacronym{mle}{MLE}{Maximun Likelihood Estimation}
\newacronym{sde}{SDE}{Stochastic Differential Equation}
\newacronym{ode}{ODE}{Ordinary Differential Equation}

\newacronym{probs}{ProbS}{Probabilistic Segmentation}
\newacronym{crf}{CRF}{Conditional Random Fields}
\newacronym{ppca}{PPCA}{Probabilistic Principal Component Analysis}
\newacronym{gmcc}{GMCC}{Generalized Multiple Correlation Coeficcient}
\newacronym{hri}{HRI}{Human-Robot Interaction}
\newacronym{ip}{IP}{Interaction Primitives}
\newacronym{hmm}{HMM}{Hidden Markov Model}
\newacronym{cac}{CAC}{Canonical Correlation Coefficient}
\newacronym{rv}{$R_v$}{$R_v$ Coefficient}
\newacronym{dcor}{dCor}{Distance Correlation}
\newacronym{dtw}{DTW}{Dynamic Time Warping}
\newacronym{edr}{EDR}{Edit Distance With Real Penalty}
\newacronym{twed}{TWED}{Time Warp Edit Distance}
\newacronym{r2}{$R^2$}{Coefficient of Determination}
\newacronym{sqp}{SQP}{Successive Quadratic Programming}
\newacronym{rkhs}{RKHS}{Reproducing Kernel Hilbert Space}
\newacronym{icnn}{ICNN}{Input-Convex Neural Network}

\newacronym{pca}{PCA}{Principal Component Analysis}

\newacronym{maf}{MAF}{Masked Autoregressive Flow}
\newacronym{iaf}{IAF}{Inverse Autoregressive Flow}
\newacronym{node}{N-ODE}{Neural ODE}
\newacronym{nsflow}{NSF}{Neural Spline Flows}
\newacronym{cnf}{CNF}{Continuous Normalizing Flows}
\newacronym{ffjord}{FFJORD}{Free-form Jacobian of Reversible Dynamics}

\newacronym{inn}{INN}{Invertible Neural Networks}
\newacronym{mcmc}{MCMC}{Markov Chain Monte Carlo}
\newacronym{ld}{LD}{Langevin Dynamics}

\newacronym{cd}{CD}{Contrastive Divergence}
\newacronym{nce}{NCE}{Noise Contrastive Estimation}
\newacronym{ce}{CE}{Cross-Entropy}
\newacronym{dsm}{DSM}{Denoising Score Matching}

\newacronym{ik}{IK}{Inverse Kinematics}

\newacronym{sdf}{SDF}{Signed Distance Field}
\newacronym{deepsdf}{DeepSDF}{Deep Signed Distace Field}

\newacronym{emd}{EMD}{Earth Mover Distance}

\newacronym{mlp}{MLP}{Multi Layer Perceptron}
\newacronym{rrt}{RRT}{Rapidly-exploring Random Trees}

\newacronym{relu}{ReLU}{Rectified Linear Unit}

\newacronym{vn}{VN}{Vector Neuron}

\newacronym{icp}{ICP}{Iterative Closest Point}

\newacronym{rmp}{RMP}{Riemannian Motion Policies}
\newacronym{mpc}{MPC}{Model Predictive Control}

\newacronym{svi}{SVI}{Structured Variational Inference}
\newacronym{vi}{VI}{Variational Inference}
\newacronym{hrl}{HRL}{Hierarchical Reinforcement Learning}
\newacronym{apf}{APF}{Artificial Potential Fields}

\newacronym{reps}{REPS}{Relative Entropy Policy Search}
\newacronym{rl}{RL}{Reinforcement Learning}

\newacronym{lmdp}{LMDP}{linearly-solvable Markov Decision Processes}
\newacronym{dwa}{DWA}{Dynamic Window Approach}
\newacronym{svf}{SVF}{Stable Vector Fields}

\newacronym{mse}{MSE}{Mean Squared Error}

\newacronym{stomp}{STOMP}{Stochastic Trajectory Optimization for Motion Planning}
\newacronym{gpmp}{GPMP}{Gaussian Process Motion Planning}
\newacronym{chomp}{CHOMP}{Covariant Hamiltonian Optimization for Motion Planning}

\newacronym{llm}{LLM}{Large Language Models}

\def\1{\bm{1}}

\def\RR{\mathbb{R}}
\def\d{{\textrm{d}}}

\def\E{{\mathbb{E}}}

\newcommand{\norm}[1]{\left\lVert#1\right\rVert}

\def\vzero{{\bm{0}}}

\def\vtheta{{\bm{\theta}}}
\def\va{{\bm{a}}}

\def\vc{{\bm{c}}}

\def\vf{{\bm{f}}}

\def\vk{{\bm{k}}}

\def\vo{{\bm{o}}}
\def\vp{{\bm{p}}}
\def\vq{{\bm{q}}}
\def\vr{{\bm{r}}}

\def\vu{{\bm{u}}}
\def\vv{{\bm{v}}}

\def\vz{{\bm{z}}}

\def\vphi{{\boldsymbol{\phi}}}

\def\vtheta{{\boldsymbol{\theta}}}

\def\mA{{\bm{A}}}

\def\mF{{\bm{F}}}

\def\mI{{\bm{I}}}

\def\mO{{\bm{O}}}

\def\mT{{\bm{T}}}

\DeclareMathAlphabet{\mathsfit}{\encodingdefault}{\sfdefault}{m}{sl}
\SetMathAlphabet{\mathsfit}{bold}{\encodingdefault}{\sfdefault}{bx}{n}

\def\gD{{\mathcal{D}}}

\def\gL{{\mathcal{L}}}

\def\gN{{\mathcal{N}}}

\def\gU{{\mathcal{U}}}

\let\oldtwocolumn\twocolumn
\renewcommand\twocolumn[1][]{%
    \oldtwocolumn[{#1}{
        \vspace{-0.7cm}
    \begin{center}
           \includegraphics[width=.9\textwidth]{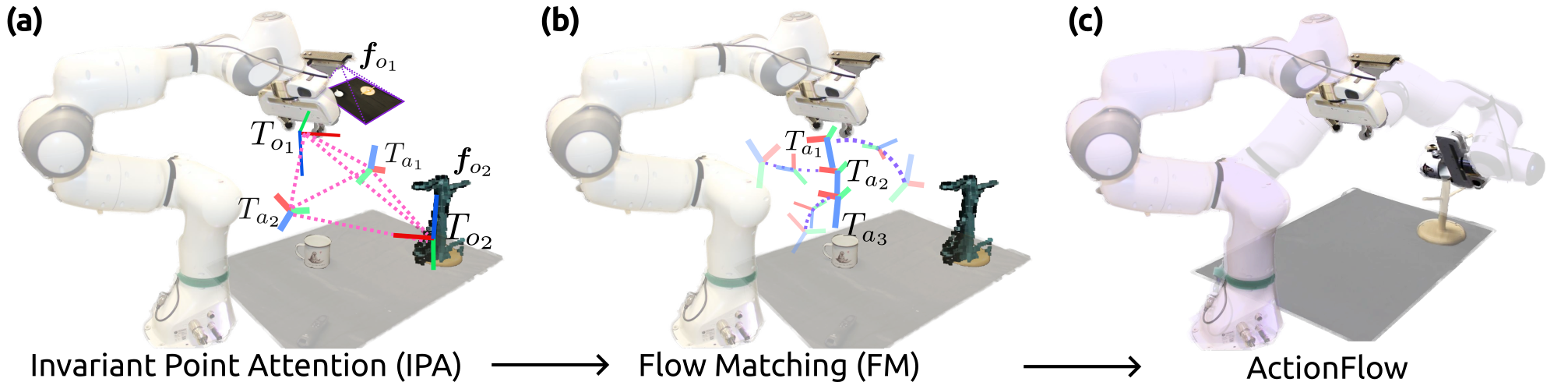}
           \vspace{-0.cm}
               \captionof{figure}{{\textbf{Overview of ActionFlow.} ActionFlow represents observations and actions in one common space and describes every token by its pose $T$ \& features $\vf$. 
    \textbf{(a)} The scene consists of an RGB and point-cloud observation ($\vf_{o1,o2}, T_{o1,o2}$), and a randomly initialized action sequence ($\vf_{a1...a3}, T_{a1...a3}$).
    Given the scene, an Invariant Point Attention-based transformer computes the attention between the tokens considering their relative SE(3) poses, thereby exploiting the task's spatial symmetries. 
    \textbf{(b)} The model's output defines a flow for refining the actions to obtain local trajectories for fulfilling the task. 
    \textbf{(c)} Iteratively using the procedure of scene encoding \& action sequence generation through Flow Matching, i.e., as a policy, generates accurate and SE(3) equivariant action sequences at low inference times.}}
    \label{fig:main}
        \end{center}
    }]
}

\let\ACMmaketitle=\maketitle
\renewcommand{\maketitle}{\begingroup\let\footnote=\thanks \ACMmaketitle\endgroup}

\makeatletter
\newcommand*\titleheader[1]{\begingroup\gdef\@titleheader{#1}\let\footnote=\thanks\endgroup}
\AtBeginDocument{%
  \let\st@red@title\@title
  \def\@title{%
  \begin{flushleft}
    \vspace{-2.15em}
    \bgroup\normalfont\small\@titleheader\par\egroup
    \vspace{-18pt}\par\noindent\rule{\textwidth}{0.1pt}
    \end{flushleft}
    \vskip0.7em\st@red@title
        }

}

\makeatother

\title{\LARGE \bf
\textsc{ActionFlow}: Equivariant, Accurate, and Efficient Policies with Spatially Symmetric Flow Matching
}

\titleheader{\textit{Preprint}}

\begin{document}

\author{Niklas Funk$^{1}$,
Julen Urain$^{1,2}$,
Joao Carvalho$^{1}$, 
Vignesh Prasad$^{1}$,
Georgia Chalvatzaki$^{1,3}$, and
Jan Peters$^{1,2,3}$
\thanks{Corresponding author: Niklas Funk. Email: niklas@robot-learning.de}%
\thanks{This work has received funding from the German Research Foundation (DFG) Emmy Noether Programme (CH 2676/1-1), the EU’s Horizon Europe projects \enquote{MANiBOT} (Grant no.: 101120823) and \enquote{ARISE} (Grant no.: 101135959), the AICO grant by the Nexplore/Hochtief Collaboration with TU Darmstadt and from the German Federal Ministry of Education and Research (BMBF) project IKIDA (01IS20045).}%
\thanks{$^{1}$Computer Science Dept., TU Darmstadt, Germany \quad $^{2}$German Research Center for AI (DFKI), Research Department: SAIROL, Darmstadt, Germany \quad $^{3}$Hessian.AI, Darmstadt, Germany
}%
\thanks{This work has been submitted to the IEEE for possible publication. Copyright may be transferred without notice, after which this version may no longer be accessible.}%
}

\maketitle

\begin{abstract}
Spatial understanding is a critical aspect of most robotic tasks, particularly when generalization is important.
Despite the impressive results of deep generative models in complex manipulation tasks, the absence of a representation that encodes intricate spatial relationships between observations and actions often limits spatial generalization, necessitating large amounts of demonstrations.
To tackle this problem, we introduce a novel policy class, \textbf{ActionFlow}. ActionFlow integrates spatial symmetry inductive biases while generating expressive action sequences.
On the representation level, ActionFlow introduces an SE(3) Invariant Transformer architecture, which enables informed spatial reasoning based on the relative SE(3) poses between observations and actions.
For action generation, ActionFlow leverages Flow Matching, a state-of-the-art deep generative model known for generating high-quality samples with fast inference -- an essential property for feedback control.
In combination, ActionFlow policies exhibit strong spatial and locality biases and SE(3)-equivariant action generation.
Our experiments demonstrate the effectiveness of ActionFlow and its two main components on several simulated and real-world robotic manipulation tasks and confirm that we can obtain equivariant, accurate, and efficient policies with spatially symmetric flow matching. 
Project website: \href{https://flowbasedpolicies.github.io/}{https://flowbasedpolicies.github.io/}
\\
\textit{Index Terms}--Learning from Demonstration, Deep Learning Methods, Deep Learning in Grasping and Manipulation
\end{abstract}

\section{Introduction}
\label{sec:introduction}
In recent years, deep generative models have demonstrated impressive results when applied as policies for solving complex manipulation tasks~\cite{zhu2023viola, chi2023diffusion, brohan2022rt, zhao2023learning}. 
However, it is well known that models that naively integrate observations and actions often require large amounts of demonstrations to achieve satisfactory performance.
In this direction, there has been a collection of research that explored how to exploit the spatial relations between observations and actions \cite{zeng2021transporter, gervet2023act3d, goyal2023rvt, vosylius2024render, shridhar2023perceiver, james2022coarse} to learn more sample-efficient policies. 
\textbf{Equivariant} policies generalize the policy's behavior under global translations or rotations in the scene~\cite{yang2023equivact, gervet2023act3d, zeng2021transporter, ryu2023diffusion, huang2024leveraging, gao2024riemann, simeonov2022neural, urain2023se}.
If the observations are rotated or translated, the generated actions will be equally transformed, thereby adding an effective inductive bias.

In this work, we are not only interested in equivariant policies that adapt to global transformations but also in \textbf{local spatial relations}~\cite{calinon2016tutorial, dreher2019learning, gervet2023act3d}.
Consider, for example, the task of picking a mug and hanging it (cf. \Cref{fig:main}). 
When the robot is approaching to pick up the mug, it should be capable of reasoning based on the relative poses between its own pose, the mug, and its next actions. But when hanging the mug, the robot should also focus on the relative poses between the mug and the hanger.
Thus, equipping the policy with the ability to reason based on the relative poses between the different observations and actions, i.e., their spatial relations, may be crucial for learning policies more efficiently.
How can we integrate all these desiderata and still learn \textbf{dexterous, fast, and expressive policies from demonstrations}?

Inspired by the recent successes from the protein folding community~\cite{jumper2021highly, yim2023fast, yim2023se, jing2024alphafold, huguet2024sequence}, in which SE(3) symmetric models are integrated with highly-expressive deep generative models, we introduce \textbf{ActionFlow}, a novel policy class for robotics, suitable for learning dexterous manipulation skills while integrating geometric notions for sample efficient learning.
In essence, ActionFlow is composed of two main elements: \textbf{(1)} a state-of-the-art~\cite{esser2024scaling} highly-expressive deep generative model (Flow Matching)~\cite{lipman2022flow, chenflow, chen2023riemannian} that has been shown capable of generating high-quality samples within very small inference times, and, \textbf{(2)} an SE(3) Invariant Transformer network~\cite{jumper2021highly, yim2023se} that frames a \textit{relative positional encoding}~\cite{jumper2021highly,liutkus2021relative} based on the tokens' relative SE(3) poses (\Cref{fig:main}).
Combining those components results in several interesting properties that make ActionFlow an appealing candidate for learning robot policies and, in particular, robotic manipulation from demonstrations:
\\
\textbf{Fast and accurate action sequence generation.} Given the Flow Matching generative model, we can generate precise action trajectories with very low inference times~\cite{liu2022flow, esser2024scaling} and run ActionFlow as an online policy.%
\\
\textbf{SE(3) equivariant action generation.} ActionFlow inherently preserves the tasks' spatial structure and naturally adapts predicted actions to corresponding observations. 
When the observations undergo translation or rotation, the actions are equally transformed, thereby providing SE(3) equivariant generation.
Although the underlying transformer network of ActionFlow is invariant, we achieve global SE(3) equivariance by applying flow matching updates relative to the actions' local frame~\cite{jumper2021highly, yim2023se}.
\\
\textbf{Relative Pose-Aware Attention.} 
The SE(3) Invariant Transformer enables actions to attend to different observation tokens based on their relative poses. This allows the system to find correlations based on the relative spatial information between the tokens, improving the generalization to scenes where objects are arranged differently.

In summary, our main contributions are: \textbf{(a)} we investigate Flow Matching for fast and precise robotic action generation, \textbf{(b)} we introduce an SE(3) Invariant Transformer architecture for geometry-aware robot learning.
Our experiments in simulated and real robot environments underline the effectiveness of both components and showcase that their combination, i.e., our proposed ActionFlow, yields accurate and fast manipulation policies while showcasing sample efficiency.

\section{Background - Flow Matching}
\label{sec:preliminaries}
Let us consider a data point $\va\in \RR^d$ and a probability path $\rho_t(\va)$ that connects a noise distribution $\rho_0(\va)$ at $t{=}0$ to the data distribution $\rho_1(\va)$ at time $t{=}1$ with its associated flow $\va_t = \vphi_t(\va_0)$, which defines the motion for the particle $\va_0$. Flow Matching~\cite{lipman2022flow} proposes learning \gls{cnf}~\cite{chen2018neural} by regressing the vector field $\vu_t(\va)=\d \vphi(\va) /\d t$ with a parametric one $\vv_{\vtheta}(\va, t)$.
In general, there is no closed-form solution for $\vu_t$ that generates $\rho_t$, making direct flow matching intractable.
Instead, \gls{cfm} proposes an efficient approach to learn \gls{cnf} by regressing a conditioned vector field $\vu_t(\va|\vz)$ that generates the probability path $\rho_t(\va|\vz)$
\begin{align}
        \gL_{\text{CFM}}(\vtheta) = \E_{t,\rho_t(\va|\vz), \rho_{\gD}(\vz)}\norm{\vv_{\vtheta}(\va,t) - \vu_t(\va|\vz)}^2,
    \label{eq:cfm}
\end{align}
with $\rho_{\gD}(\vz)$ being the data distribution. 
As shown in \cite{lipman2022flow}, 
$\vv_{\vtheta}$ recovers the marginalized conditioned vector field
$\vu_t = \int_\vz \vu_t(\va|\vz) \rho_t(\va|\vz) \rho_{\gD}(\vz)/\rho_t(\va) \d \vz$
that generates the marginalized distribution path $\rho_t = \int_\vz \rho_t(\va|\vz) \rho_{\gD}(\vz) \d \vz$.
Then, the problem boils down to designing a conditioned vector field $\vu_t(\va|\vz)$ that 
moves a randomly sampled point at time $t=0$ to the datapoint $\vz$ at time $t=1$.

\section{ActionFlow}
\label{sec:flow_se3}
The desiderata for ActionFlow policies is to be fast, accurate, expressive, and sample-efficient.
Particularly, ActionFlow should capture the spatial relations between observations and actions and yield SE(3) equivariant action generation.
To achieve these properties, ActionFlow is built on two core elements: a Flow Matching-based generative model that generates action sequences quickly, and a geometrically grounded transformer model capable of capturing the intricate spatial relations between observations and actions in the SE(3) space.

Before proceeding with presenting both components in detail, we introduce the structure of ActionFlow's observation and action space.
Generally speaking, ActionFlow relies on a geometrically grounded scene representation.
Both the observations $\mO:(\mT_o,\mF_o)$ and actions $\mA:(\mT_a,\mF_a)$ are represented through a sequence of poses $\mT = (T^1,\dots, T^N)$ and features $\mF = (\vf^1,\dots, \vf^N)$ as shown in \Cref{fig:main}.
In other words, every individual action or observation consists of a pose $T = (\vr, \vp)\in SE(3)$, with rotation matrix $\vr \in SO(3)$ and 3D position vector $\vp \in \RR^3$, together with an associated feature $\vf \in \RR^d$ representing semantic information related to the specific action/observation.
This representation is highly flexible.
For the task of placing a mug onto a hanger as shown in \Cref{fig:main} - left, the RGB image from the wrist-mounted camera is represented by the camera's current pose, i.e., the location from which the image was captured, while the features correspond to the encoded image observation.
Given that the wrist view provides a very localized observation, to solve the task, we additionally provide as input to the policy the point cloud observation of the hanger which is obtained from an initial global view of the scene.
The point cloud features are obtained through a point cloud encoder, while the hanger's pose is centered at the mean of the point cloud.
In other environments, where, e.g., an explicit object pose estimator \cite{wen2023foundationpose} can be deployed, the objects' poses are given by the output of the pose estimator, while the features can represent semantic information that describes that object (color, shape, \dots).
An action pose represents the desired target pose that should be reached, while the features represent at what instant of time this target pose should be reached.
This information is crucial since we are predicting a sequence of actions.
Given this representation, our goal is to learn ActionFlow policies $\pi_{\vtheta}(\mT_a|\mO)$ that generate action pose sequences $\mT_a$ given an observation $\mO$. 

\subsection{Flow Matching for SE(3) Action Generation}
Flow Matching permits the learning of expressive generative models with fast inference.
Similarly to diffusion models~\cite{song2020score, ho2020denoising}, sample generation is done by iteratively refining initially noisy samples.
However, as shown in \cite{liu2022flow, esser2024scaling}, Flow Matching-based models require fewer calls to the model, reducing the inference time of generating samples.
In our work, we apply (conditioned) Flow Matching to generate action pose sequences.
We consider an action space represented with a sequence of $N$ SE(3) poses $\mT_a = (T^1_a,\dots, T^N_a) \in SE(3)^N$.
Thus, similarly to \cite{yim2023fast}, we adapt Flow Matching to the Lie Group SE(3).
Without loss of generality, we derive the solution for a single SE(3) action pose $T\in SE(3)$, yet the same derivation holds for generating a sequence of actions.
For completeness, we also derive Flow Matching for action generation in the Euclidean space in \Cref{app:flow_pi}.

We adapt a common Flow Matching method (Rectified Linear Flow~\cite{liu2022flow, esser2024scaling, chenflow}) to the Lie Group SE(3).
We define a decoupled flow between the rotation and the translation, allowing us to represent the distribution path and the vector fields independently.
\\
\textbf{SE(3)-Rectified Linear Flow.} 
The Rectified Linear Flow proposes representing the datapoint-conditioned flow $\vphi_t(\va|\va_1)$ with a straight line from a noisy sample $\va_0{\sim}\gN(\vzero,\mI)$ at $t{=}0$ to the datapoint $\va_1\in\gD$ at $t{=}1$.
In our case, we want to move an initially randomly sampled action pose $T_0$ towards an action pose sampled from the dataset $T_1$ by defining a straight line path for both the translation $\vp$ and the rotation $\vr$ component of the SE(3) pose.
Specifically, the flow $T_t = \vphi_t(T_0|T_1)$ that moves a noisy initial sample $(\vp_0,\vr_0)$ to a pose sampled from the dataset, i.e., $(\vp_1,\vr_1)\sim \gD$, is represented by
\begin{align}
\begin{split}
    \text{Translation:~} \vp_t &= \vphi_t(\vp_0|\vp_1) = t\vp_1 + (1-t)\vp_0 \\
    \text{Rotation:~} \vr_t &= \vphi_t(\vr_0|\vr_1) = \vr_0\text{Exp}\left(t \text{Log}(\vr_0^{-1}\vr_1)\right),
\end{split}
\label{eq:rlf_so3}
\end{align}
with $\text{Log}$ and $\text{Exp}$ being the logarithmic and the exponential map of the SO(3) manifold, respectively~\cite{sola2018micro}.
Notice that \Cref{eq:rlf_so3} represents a path through the geodesic on SO(3) from $\vr_0$ to $\vr_1$.
\begin{figure*}[t]
    \centering
    \makebox[\textwidth][c]%
    {\includegraphics[width=.9\textwidth]{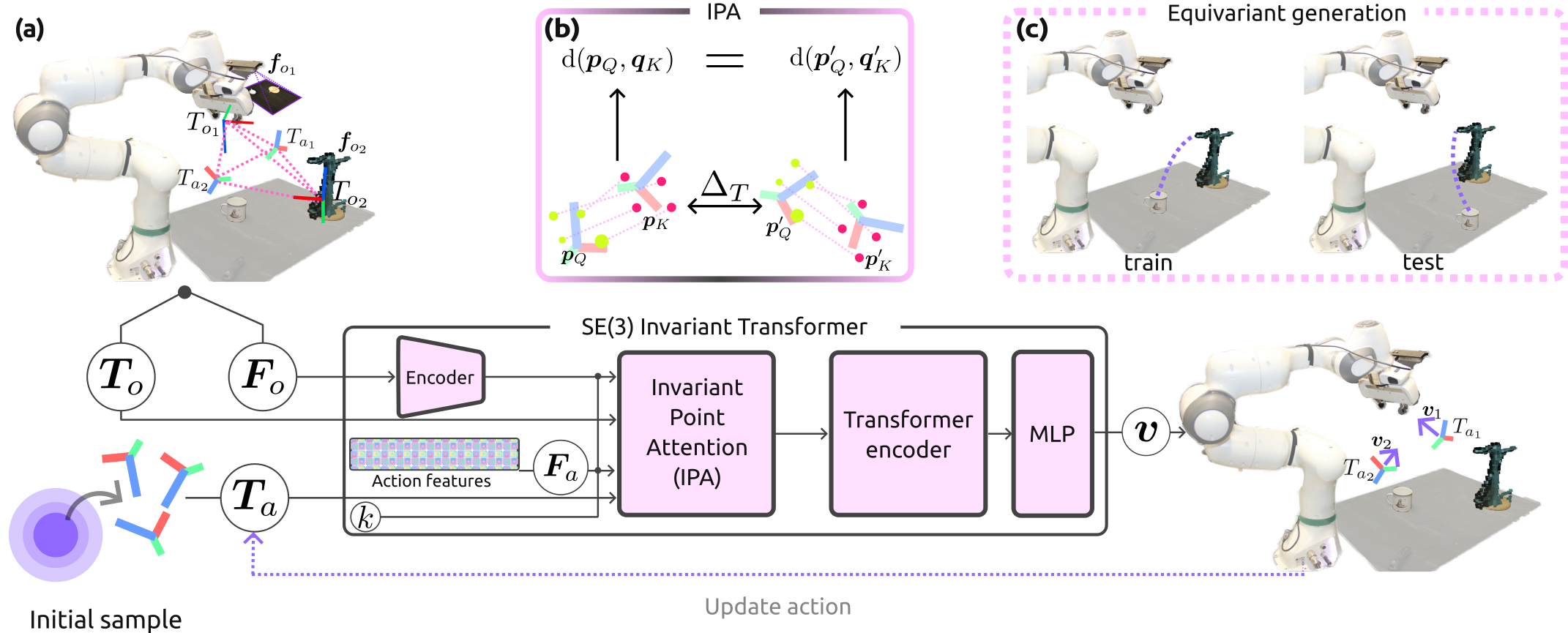}}   
    \caption{{\textbf{Spatial Symmetries in ActionFlow}. \textbf{(a)}, Visual representation of the SE(3) Invariant Transformer. Given a set of observations $\bm{\mF_o}$ with associated poses $\mT_o$ \& candidate actions $\mT_a$, the Transformer predicts vectors $\vv$ to update the actions (\Cref{eq:sample_se3}). The action refinement process is repeated $K$ times. \textbf{(b)} \gls{ipa} augments the classical attention with points $\vp_Q$ and $\vp_K$ generated in the local frames of the query and key poses. The layer is designed to generate the same output under global SE(3) transformations $\Delta_T \in SE(3)$. \textbf{(c)} ActionFlow is SE(3) equivariant. If we apply a transformation over the observation poses, the generated actions will be equally transformed (cf. \Cref{app:equiv_gen}).}}
    \label{fig:se3_invariant_transformer}
\vspace{-0.5cm}
\end{figure*}
Given the flow is decoupled, the vector field 
$\vu_t = \d \vphi_t / \d t$
is also decoupled. In particular, the translation velocity $\dot{\vp}_t\in\RR^3$ and the rotation velocity $\dot{\vr}_t\in\RR^3$ equate to
\begin{align}
    \dot{\vp}_t = \vr_t^{\intercal} \frac{\vp_t - \vp_1}{1-t} \quad ,\quad  \dot{\vr}_t = \frac{\text{Log}(\vr_t^{-1}\vr_1)}{1-t}.
    \label{eq:se3rlf}
\end{align}
Notice that even if rotations are represented in $\vr\in SO(3)$, the velocity vector for the rotations $\dot{\vr}_t\in \RR^3$ is a 3D vector (axis-angle representation) represented in the tangent space centered around $\vr_t$.
We also point out that the translation velocity is premultiplied with the transpose of the current rotation $\vr_t^{\intercal}$ as we aim to represent the velocity in the action's local frame.
\\
\textbf{Training.}
Our parameterized Flow Matching model $(\vv_{\vp}, \vv_{\vr}) = \vv_{\vtheta}(T,\mO, t)$ should output both a translation vector ${\vv_{\vp}\in\RR^3}$ and a rotation vector ${\vv_{\vr}\in\RR^3}$ for refining the sample.
Similar to \Cref{eq:cfm}, given a dataset $\gD:\{T^i, \mO^i\}_{i=0}^I$, the training objective is to minimize the mean-squared error between the model outputs $(\vv_{\vp}, \vv_{\vr})$ and the velocity vectors $\vu_t=(\dot{\vp}_t, \dot{\vr}_t)$ from \Cref{eq:se3rlf}, i.e., $\gL = || \vv_p - \dot{\vp}_t ||^2 + || \vv_r - \dot{\vr}_t ||^2$.
We add additional details in \Cref{app:se3_fm}.
\\
\textbf{Sampling in SE(3).} To generate an action pose $T=(\vp,\vr)$, we initially sample a random rotation $\vr_0 \sim \gU(SO(3))$ and translation $\vp_0\sim \gN(\vzero,\mI)$ and 
iteratively run Euler-discretization for $K$ inference steps to refine the sample
\begin{align}
\begin{split}
    \vp_{k+1} &= \vp_k + \vr_k\vv_{\vtheta}(T_k,\mO, t)\Delta t \\
    \vr_{k+1} &= \vr_k \text{Exp}(\Delta t \vv_{\vtheta}(T_k,\mO, t))
    \label{eq:sample_se3}
\end{split}
\end{align}
with $\Delta t {=} 1/K$ and $t{=}k\Delta t$. 
Notice that since the translation velocity is also computed in the current pose frame $T_k$, it is first transformed to the world frame given the relative rotation $\vr_k$ before applying the Euler discretization.

\glsreset{ipa}

\subsection{SE(3) Invariant Transformer}
\label{sec:se3_inv_trans}
As model architecture of ActionFlow, we propose an SE(3) Invariant Transformer~(cf. \Cref{fig:se3_invariant_transformer}).
At the core of this architecture is a geometry-aware attention layer, known as \gls{ipa}~\cite{jumper2021highly, yim2023se}.
The \gls{ipa} layer augments the queries, keys, and values of classical attention~\cite{vaswani2017attention} with a set of 3D points that are generated in the local frames of the
query $\mT_Q$ and key $\mT_K$ poses. 
The layer is designed such that the output is invariant to global rotations and translations (cf. \Cref{fig:se3_invariant_transformer} (b)).
If we apply a transformation $\Delta_T \in SE(3)$ over both observation poses $\mT_o' {=} \Delta_T \mT_o$ and action poses $\mT_a' {=} \Delta_T \mT_a$, the network generates the same output.
Moreover, with the \gls{ipa} layer, the network can reason about all the relative poses between the entities in the scene.
We hypothesize that the invariant and object-centric nature of the network will lead to more data-efficient policies.

\textbf{Network Architecture.} Given an observation $\mO = (\mT_o, \mF_o)$ and a candidate action sequence $\mT_a \in SE(3)^N$ of length $N$, the SE(3) Invariant Transformer outputs vectors $\vv\in\RR^{6\times N}$ that predict the direction in which the actions $\mT_a$ should be updated following \Cref{eq:sample_se3}.
Our network architecture is inspired by the protein folding network FrameDiff~\cite{yim2023se, yim2023fast} that combines an \gls{ipa} layer with a transformer encoder.
Given a set of poses $\mT$ and features $\mF$, the network first applies an \gls{ipa} layer to capture the spatial relative attention between the different entities, followed by a transformer encoder to find higher-order interactions. 
We use a small linear layer to map the transformer output to the vector $\vv$.
Notice that prior to the \gls{ipa}, we employ an observation encoder that maps all observation features onto a common latent feature space. The action features $\mF_A$ are given by a learnable parameter vector.
\\
\textbf{Action Generation.} For action generation, we start with a randomly sampled action sequence $\mT_a$ and iteratively update the actions (cf. \Cref{eq:sample_se3}) by calling the SE(3) invariant transformer $K$ times.
Given the invariant network and the action updates within their current local frame, the resulting policy $\pi_{\vtheta}(\mT_a|\mT_o,\mF_o)$ is equivariant. %
If we apply a transformation over $\mT_o$, the generated actions $\mT_a$ are guaranteed to be equally transformed (cf. \Cref{fig:se3_invariant_transformer} (c) \cite{jumper2021highly}.
We provide further insights about equivariant action generation with invariant models in \Cref{app:equiv_gen}.

\section{Experimental Results}
\label{sec:result}
\begin{figure*}[t]
    \centering
    \includegraphics[width=.99\textwidth]{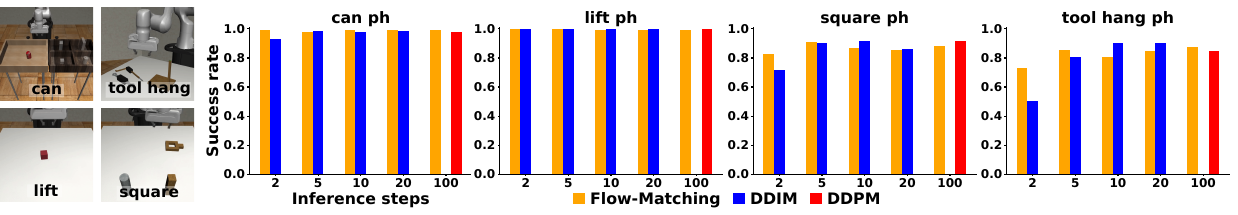}
    \caption{ 
    \textbf{Success rate evaluation on Robomimic tasks} with state-based observations averaged over 3 seeds and 50 environments initializations.
    We evaluate flow-matching and diffusion policy~\cite{chi2023diffusion} with different inference steps. For diffusion policy, we use DDPM for $100$ inference steps and DDIM otherwise.
    }
    \label{fig:robomimic_exps_final}
\vspace{-0.5cm}
\end{figure*}

The experiments are divided into three parts. 
First, we explore the performance of flow matching in generating high-quality samples in small inference steps.
Second, we evaluate the impact of the SE(3) Invariant Transformer, particularly, whether \gls{ipa} helps in learning policies in a data-efficient manner.
Third, we evaluate ActionFlow in two real robot manipulation tasks.\footnote{Experiment videos and appendix with details are \href{https://flowbasedpolicies.github.io/}{on our website.}}

\subsection{Flow Matching for Fast and Accurate Action Generation}
\label{sec:exp_1}

We compare Flow Matching against Diffusion Policy~\cite{chi2023diffusion} for action sequence generation in the simulated Robomimic tasks~\cite{mandlekar2022matters}.
To ensure a fair comparison, both methods use the same transformer architecture from~\cite{chi2023diffusion} to either model the (observation-conditioned) vector field in \Cref{eq:cfm} or the denoising network~\cite{chi2023diffusion}.
Moreover, we consider Flow Matching in the Euclidean space (cf. App. \ref{app:flow_pi}) and use the same hyperparameters from the Diffusion Policy in Flow Matching.
Both policies are conditioned on the current observations and trained for $4000$ epochs with $K{=}100$ time steps. Checkpoints are evaluated every $200$ episodes.
For testing, we pick the best-performing checkpoint during training and report the average success rate from policy rollouts starting from $50$ different initial configurations (from the test set) across $3$ training seeds.
During inference, it is desirable to use fewer steps than during training since it enables higher-frequency policies.
In Flow Matching, reduced inference time steps are obtained by interpolating the training time steps, while for Diffusion Policy we use DDIM~\cite{song2021ddim} for faster sampling.
For more information, see App. \ref{app:robomimic_experiments}.

\textbf{Results.}
\Cref{fig:robomimic_exps_final} shows the results.
We depict the success rates for each method when varying the number of available inference steps.
Since both methods use the same underlying transformer network, they require the same amount of time per inference step.
In practice, $\{2, 5, 10, 20, 100\}$ inference steps allow for action sequence generation with the respective policy at $\{100, 33, 20, 9, 2\}$~Hz, respectively.
Therefore, fewer inference steps are desirable as they either allow for faster action generation, or for reducing the required computations, i.e., the overall number of action refinement steps.
As shown, for most environments and available inference steps, Flow Matching and diffusion perform almost equally.
However, Flow Matching results surpass those from Diffusion Policy for very small inference steps.
This effect is most noticeable in the Tool Hang task, which requires the policy to produce very accurate actions.
These experiments show that Flow Matching policies can maintain good success rates when using only $2$ inference steps (allowing for action generation at \SI{100}{\hertz}) while Diffusion Policy results degrade.

\subsection{SE(3) Invariant Transformer Evaluation: Multi-Token Observations and Invariant Point Attention}
\label{sec:exp_2}

\begin{figure}
\begin{center}
\begin{minipage}{\columnwidth}
    \centering
    \includegraphics[width=.99\textwidth]{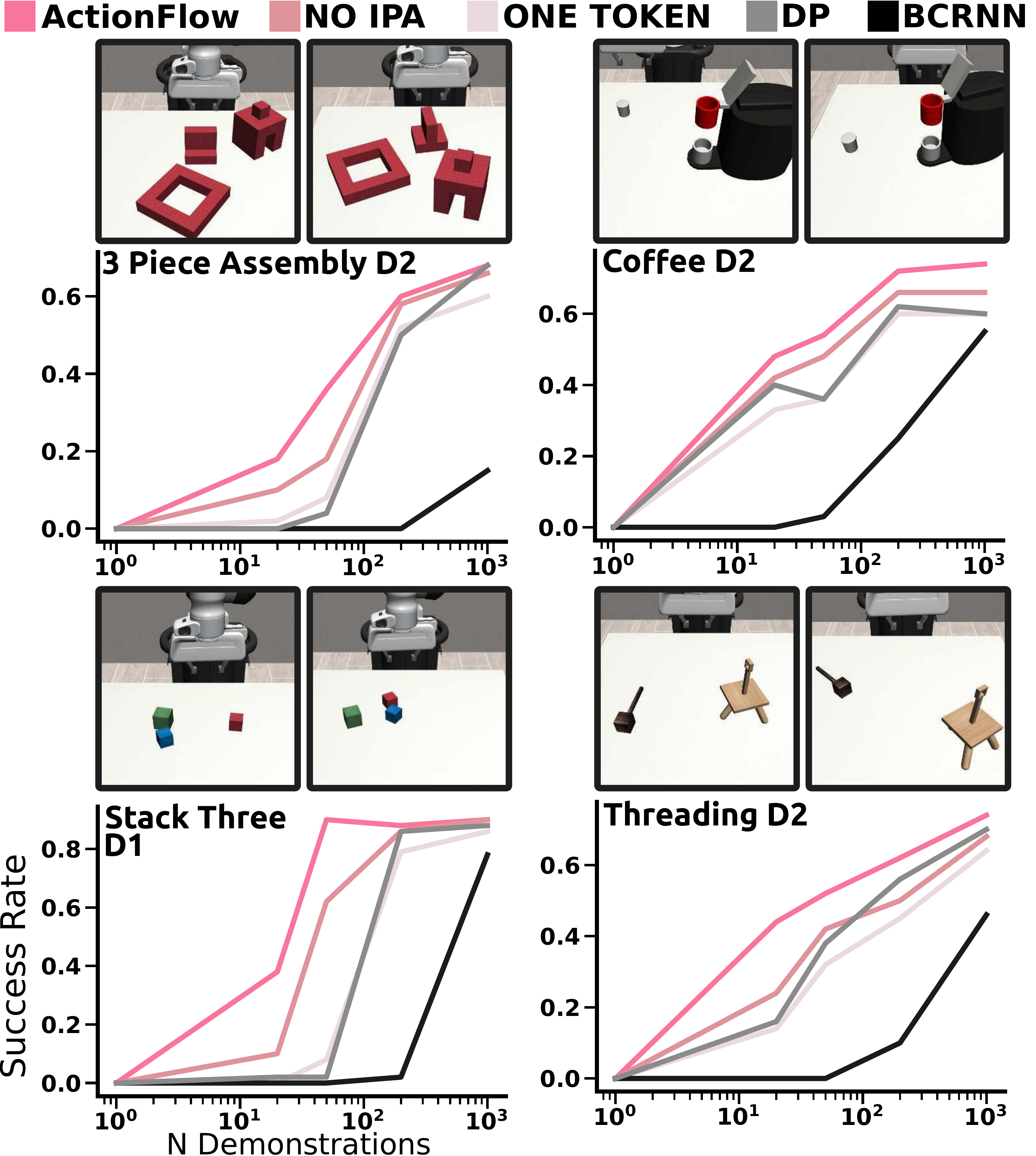}
\end{minipage}
\end{center}
\vspace{-0.3cm}
\caption{\textbf{Success rate of models trained on different number of demonstrations} $(20, 50, 200, 1000)$ on a subset of Mimicgen tasks~\cite{mandlekar2023mimicgen}.
We report the best success rate obtained by the model configurations on 50 test environments. 
The top row shows two randomly sampled initial configurations for each task.}
\label{fig:experiment_2}
\vspace{-0.45cm}
\end{figure}

This experiment evaluates the performance of the proposed SE(3) Invariant Transformer (cf. \Cref{sec:se3_inv_trans}).
Specifically, we aim to evaluate the influence of two design choices:
\\
(\textbf{1}) Does a \textbf{multi-token representation}, in which each object is treated as a single token, enhance policy performance?
\\
(\textbf{2}) Does the \textbf{\gls{ipa}} layer, which allows computing the relative poses of all tokens among each other, help in finding informative features to improve policy performance?
\\
\textbf{Dataset \& Evaluation Environment.} 
The experiments are conducted in a subset of Mimicgen environments~\cite{mandlekar2023mimicgen}. The datasets consist of 1000 synthetically generated demonstrations, given 10 original demonstrations.
The original dataset provides the observations as a single vector and represents the action displacements in the world frame.
We slightly adapt the data to be compatible with our model ($\mT,\mF$) and represent it as object poses $T_o^w$ and action poses $T_a^w$ in the world frame (cf. \Cref{app:obs_act_rep}).
\\
\textbf{Models.} We consider three variations of ActionFlow:
(\textbf{A.1}) The original ActionFlow as introduced in \Cref{sec:se3_inv_trans}.
(\textbf{A.2}) We eliminate the \gls{ipa} layer but keep each object as an independent token.
(\textbf{A.3}) We eliminate the \gls{ipa} layer and represent all observations as a single token.
All models consider $100$ inference steps. 
Additionally, we consider as baseline (\textbf{B}) a Diffusion Policy (DP) model \cite{chi2023diffusion} ($100$ denoising steps), and (\textbf{C}) a RNN-GMM model as introduced in \cite{mandlekar2022matters, mandlekar2023mimicgen}.
\\
\textbf{Experiment \& Results.} We train the models with different amounts of demonstrations $(20, 50, 200, 1000)$ for $3500$ epochs and evaluate their performance in $50$ randomly sampled test environments. 
The results in \Cref{fig:experiment_2} reveal that the original ActionFlow consistently outperforms the variations and baselines across the tasks, specifically in low demonstration regimes.
This indicates that \gls{ipa} is beneficial for learning policies in a sample-efficient way, while with large datasets, all models appear to converge to similar performances.
We also observe that representing the observations with multiple tokens shows performance benefits compared to representing the whole observation as a single token.
Moreover, the Diffusion Policy (DP) baseline performs very similar to our model ablation A.3 with the single token.
One potential explanation for this finding is that the DP's underlying transformer model also flattens the observations of each timestep into a single token, thereby resembling a similar network architecture as the single token ablation.
Finally, all ActionFlow variations outperform the BCRNN baseline in different data percentages.
We hypothesize that this performance increase could be directly related to the expressivity of Flow Matching with respect to GMM.

\begin{figure}[t]
\centering
    \includegraphics[width=.99\columnwidth]{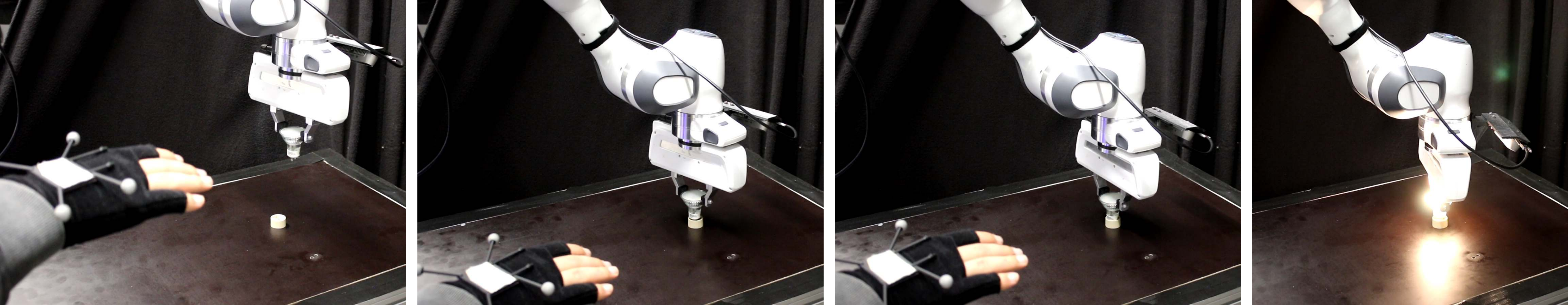}
\caption{
\textbf{Teleoperation Interface} used to collect the data. This image series depicts the data collection process for the lightbulb mounting experiment.  
}
\vspace{-0.25cm}
\label{fig:teleop_lightbulb_mounting}
\end{figure}

\subsection{Real Robot Experiments}
\label{sec:rr_exps}
We finally evaluate ActionFlow in two real robotic tasks: (i) mounting a light bulb, and (ii) placing a mug onto a hanger.
While the first task assesses ActionFlow's accuracy, the second investigates its equivariance.
\\
\textbf{Setup.} The experimental platform consists of a 7DoF Franka Panda manipulator with a RealSense mounted at its end-effector (cf. Fig. \ref{fig:main}).
For both tasks, we employ a token for the robot's end-effector $(T,\vf)$, with the end-effector pose $T$ and its features $\vf$ consisting of the encoded RGB RealSense image and the current gripper opening width.
We use an observation history of 5 steps and predict an action sequence containing 16 steps.
While the RealSense camera returns RGB readings with a resolution of $640 \times 480$, we resize the images to $80 \times 80$ pixels before passing them through the ResNet18 \cite{he2016resnet} for obtaining the encodings.
The resizing helps to reduce the dataset's size significantly and, therefore, facilitates \& speeds up policy training.
The ResNet18 for encoding the images is trained from scratch.
We parameterize our ActionFlow policies for real robot manipulation using the SE(3) Invariant Transformer with $K=4$ inference steps, train the policies for $75$ epochs, and evaluate the last checkpoint.
Our computer is equipped with an AMD EPYC 7453 CPU; 512 GB RAM; RTX 3090 Turbo GPU.
\\
\textbf{Data Collection \& Control.}
For data collection through teleoperation, we leverage an off-the-shelf presenter \cite{AmazonPresenter} for controlling, i.e., opening and closing the gripper.
To control the pose of the robot's end-effector, we rely on the OptiTrack Motion capture system.
In particular, as shown in \Cref{fig:teleop_lightbulb_mounting}, the teleoperator wears a glove that has OptiTrack markers rigidly attached to it.
Upon starting teleoperation, the glove's current pose is defined as the reference.
Moving the glove w.r.t.~this reference moves the robot's end-effector w.r.t.~its initial pose accordingly.
The teleoperation interface is set to operate at \SI{25}{\hertz}.
For controlling the robot during teleoperation and policy rollout, we leverage the Cartesian Pose Impedance Controller from \cite{Nbfigueroafranka_interactive_controllers}.
We provide further details in \Cref{app:reaL_robot}.
\begin{figure}
\centering
\begin{minipage}{.65\columnwidth}
    \resizebox{.99\textwidth}{!}{%
    \centering
    \begin{tblr}{
        colspec={l | c },
        row{1}={font=\bfseries\footnotesize},
        row{2-20}={font=\footnotesize},
        row{3-4} = {bg=gray!10},
        width=\linewidth,
    }
        \SetCell[c=1,r=1]{l} \textbf{Method} & \SetCell[c=1,r=1]{l} \textbf{Success Rate} \\
        \hline
        Replay Demonstrations & 8/10 \\
        Perturbed Demonstrations Replay & 5/10 \\
        \textbf{ActionFlow (Ours)} &  8/10 \\
    \end{tblr}
    }
\end{minipage}
\begin{minipage}{\columnwidth}
    \centering
    \includegraphics[width=.99\textwidth]{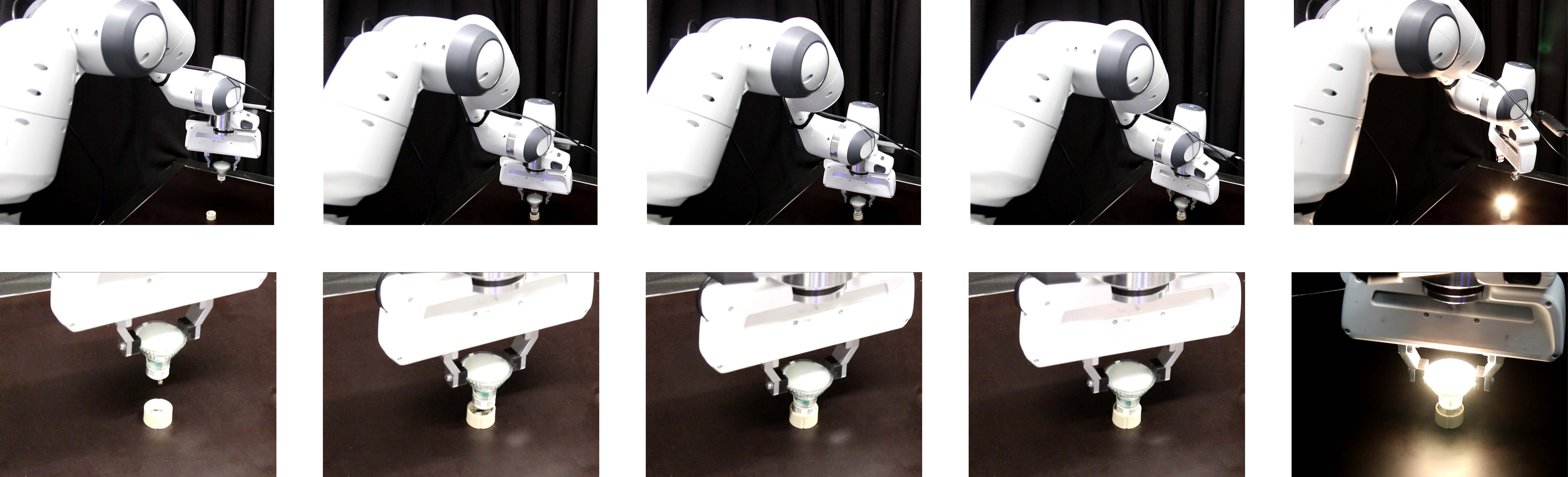}
\end{minipage}
\caption{\textbf{Real robot light bulb experiments}. \textbf{Top:} We report the performance of our model and two baselines, i.e., replaying 10 randomly selected demonstrations and replaying the demonstrations with a position offset in one randomly selected direction of \SI{1.5}{\mm}. \textbf{Bottom:} Successful rollouts illustrating that ActionFlow policies can generate highly accurate trajectories.}
\label{fig:experiment_bulbscrew}
\vspace{-0.5cm}
\end{figure}

\textbf{Lightbulb Mounting Experiment - Evaluating Accurate Action Sequence Generation.}
\\
The task of mounting a lightbulb (cf.~\Cref{fig:experiment_bulbscrew}) consists of two main phases: first, the lightbulb and its two pins have to be precisely inserted into the socket.
Second, the bulb has to be rotated to tighten it and turn it on.
For this task, we collected 100 demonstrations using our teleoperation interface.
\\
We consider two baselines. 
The first one consists of replaying 10 randomly selected demonstrations. 
The second one is replaying 10 randomly selected demonstrations with an additional offset of \SI{1.5}{\mm} sampled in a random direction.
The results in \Cref{fig:experiment_bulbscrew} show that replaying 10 randomly selected demonstrations results in 8 out of 10 successful executions.
Adding a small perturbation of \SI{1.5}{\mm} reduces the number of successes to only 5.
These findings underline that this task requires high accuracy and precise trajectory execution, as slight perturbations significantly reduce the success rate.
Despite the tight tolerances and initializing in perturbed states, our learned ActionFlow policy, which is only conditioned on the input of the RGB camera, achieves a success rate of $80\%$.
Two successful policy rollouts are presented in \Cref{fig:experiment_bulbscrew}.
This experiment underlines that ActionFlow is capable of generating highly accurate trajectories suitable for solving delicate manipulation tasks.

\begin{figure}[t]
\centering
\includegraphics[width=0.85\columnwidth]{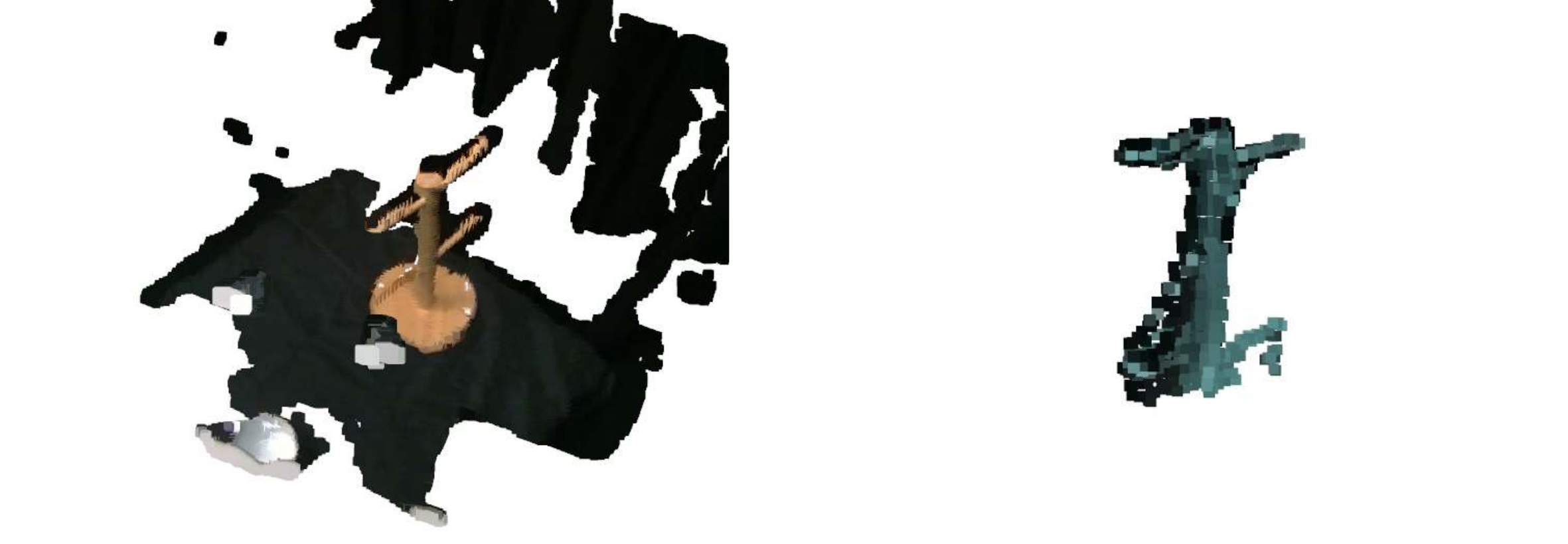}
\caption{
\textbf{}Illustrating the hanger's point cloud (right) after applying a box filtering to the initial views and adding them. One of the initial views is depicted on the left.}
\label{fig:pcl_hanger_extracted}
\begin{center}
\begin{minipage}{0.55\columnwidth}
    \centering
    \includegraphics[width=.99\textwidth]{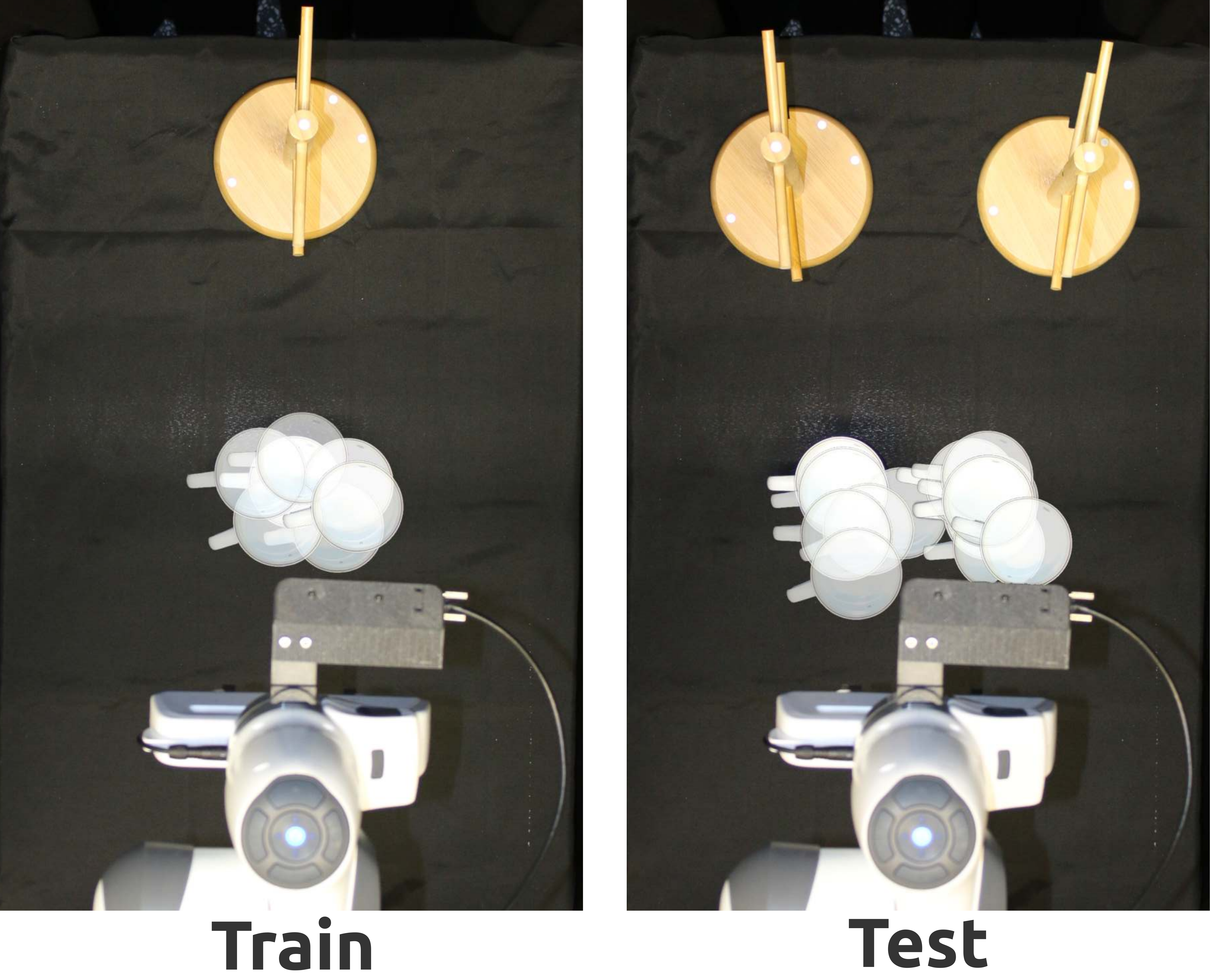}
\end{minipage}
\end{center}
\vspace{-0.3cm}
\caption{Visualization of the initialization configurations for training and testing evaluations in the mug hanging experiment.}
\label{fig:experiment_mug_hang_init}
\end{figure}

\textbf{Mug Hanging Experiment - Evaluating Equivariance.}
For the task of picking up a mug and placing it onto a hanger (cf. \Cref{fig:main}), we assume as observations the robot's proprioception (for obtaining the end-effector pose), and the information from the wrist-mounted RGB-D camera.
Therefore, the hanger's pose should be extracted from the camera's readings, particularly its depth readings.
For solving the mug hanging task solely based on proprioception and the camera's readings, before starting the teleoperation (for data collection) or the policy rollout, the robot visits seven predefined end-effector poses, which ensure good visibility of the hanger (cf. \Cref{fig:pcl_hanger_extracted} - left).
After adding these 7 point cloud measurements and applying a box filter to remove the background \& the ground plane, we are left with a point cloud representation of the hanger as shown in \Cref{fig:pcl_hanger_extracted} - right.
This point cloud is passed as an additional token into our SE(3) Invariant Transformer model.
In particular, the point cloud's features are obtained using a PointNetEncoder \cite{deng2021vector}, while the token's pose is obtained through the mean of the point cloud, while the rotation is set to identity.
For policy training, we collect 50 demonstrations using variations as
shown in the left of \Cref{fig:experiment_mug_hang_init}.
Notably, the demonstrations only include slight variations of the mug poses, while the hanger always stays in the same pose.

\begin{figure}
\centering
\begin{minipage}{.6\columnwidth}
    \resizebox{.99\textwidth}{!}{%
    \centering
    \begin{tblr}{
        colspec={l | c },
        row{1}={font=\bfseries\footnotesize},
        row{2-20}={font=\footnotesize},
        row{3-3} = {bg=gray!10},
        width=\linewidth,
    }
        \textbf{Initialization} & \textbf{ActionFlow Success Rate} \\
        \hline
        Train & 9/10 \\
        Test &  7/10 \\
    \end{tblr}
    }
\end{minipage}
\begin{minipage}{\columnwidth}
    \centering
    \includegraphics[width=.99\textwidth]{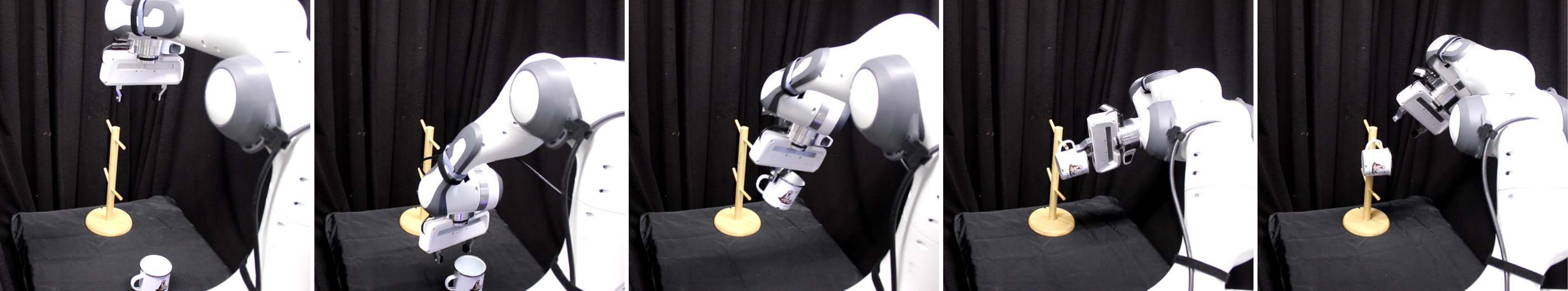}
\end{minipage}
\caption{\textbf{Top:} Results from the point cloud-conditioned mug hanging experiments. \textbf{Bottom:} Illustration of a successful ActionFlow policy rollout for this task.}
\label{fig:results_pcl_mug_main}
\vspace{-0.25cm}
\end{figure}

The results are presented in \Cref{fig:results_pcl_mug_main}.
The table's first row reveals that the ActionFlow policy achieves high success rates of 90\% upon evaluating in similar scenarios as those encountered during training.
We only observe one failure in which the mug is not grasped properly.
Importantly, our policies run online in real-time as action generation takes \SI{0.04}{\s} on an NVIDIA RTX $3090$ GPU.
These results also underline that our ActionFlow policies can handle different sensing modalities, since our transformer effectively combines the RGB view of the wrist camera and the point cloud observation providing information about the hanger's location.
We also evaluate the policy in previously unseen test scenarios (cf.~\Cref{fig:experiment_mug_hang_init}), where the hanger is moved to either side of the table.
The results show that our policy can still handle these novel test scenarios well, achieving 70\% success.
We repeated the experiment with a different ActionFlow policy which directly has the hanger's pose available through OptiTrack, i.e., it does not need to be extracted from a Point Cloud.
This variation achieves the same performance in the previously seen train configurations (9/10) and performs even slightly better in the test scenarios, with 8 out of 10 successes.
These findings might be attributed to the fact that in the real world, moving the hanger might cause a slight change in its point cloud appearance and encoding, thereby slightly reducing the performance in the unseen testing scenarios.
Nevertheless, the experiments and the good generalization to the previously unseen testing scenarios highlight the equivariance properties of our proposed ActionFlow framework.

\section{Related Work}
\label{sec:related_work}
\textbf{Exploiting Spatial Symmetries for Robot Learning.}
Despite the impressive performance of learning policies from demonstration data for dexterous robotic manipulation~\cite{chi2023diffusion, zhao2023learning, brohan2022rt}, naively combining observations and actions leads to large data requirements to achieve good performances.
A large line of research, therefore, proposed methods for better aligning observation and action spaces to alleviate the data requirements of learning from demonstrations~\cite{zeng2021transporter, gervet2023act3d, goyal2023rvt, james2022coarse, simeonov2022neural, urain2023se, ke20243d, shen2023distilled}.
In particular, the methods can be grouped into works that propose to represent and optimize the actions directly in a visual, pixel-related space~\cite{zeng2021transporter, goyal2023rvt, vosylius2024render, huang2022equivariant, shridhar2022cliport, lin2023mira}.
Alternatively, and more closely related to this work, other approaches represent both observations and actions in a three-dimensional space~\cite{gervet2023act3d, shridhar2023perceiver, yang2023equivact, simeonov2022neural, urain2023se, ke20243d, shen2023distilled, shafiullah2022clip, pan2023tax, ryu2023equivariant, huang2024fourier, ze20243d}.
However, most of the previously mentioned methods~\cite{zeng2021transporter, goyal2023rvt, shridhar2023perceiver, simeonov2022neural, urain2023se, shen2023distilled, huang2022equivariant, shridhar2022cliport, lin2023mira, pan2023tax, ryu2023equivariant, huang2024fourier} focus on solving manipulation tasks by solely optimizing for the robot's grasping and placing poses, thereby necessitating an additional planning module for obtaining the robot's motion.
Conversely, this work presents a novel policy class suitable for online, reactive motion generation in real time.
Our proposed method is capable of handling pick-and-place and fine-insertion tasks without any additional intermediate motion planning.
Moreover, inspired by \cite{yang2023equivact, huang2022equivariant, ryu2023equivariant, huang2024fourier}, our proposed method also exploits the concept of SE(3)-equivariance to further boost efficiency w.r.t. learning from few demonstrations.
Additionally, while prior works operate within specific observation spaces (e.g., point clouds \cite{yang2023equivact, ze20243d} or images \cite{vosylius2024render}), our proposed transformer architecture can seamlessly handle and combine different sensing modalities, leveraging a geometrically grounded representation in the three-dimensional space.
Lastly, \cite{gervet2023act3d, ke20243d} introduce the concept of spatial relative attention to enable the policy to reason based on the relative 3D positions between observations and actions.
Herein, we extend this concept to SE(3) poses, allowing the agent to reason based on the relative position and orientation between observations and actions.
\\
\textbf{Flow Matching for Decision Making.}
Despite the recent emergence of flow matching methods~\cite{lipman2022flow, liu2022flow, chen2023riemannian} for learning deep generative models, there has been a wide set of fields in which they have already been applied, ranging from image generation~\cite{esser2024scaling} to protein backbone generation~\cite{yim2023fast, bose2023se}.
In the context of decision making, Zheng et al.~\cite{zheng2023guided} introduce guided flow matching to condition the flow-based models on arbitrary contexts. Then, they apply the conditioned flow-based models in Offline RL setups, similarly to \cite{janner2022planning}.
Concurrent to this work, Braun et al.~\cite{braun2024riemannian} introduce Riemannian Flow Matching Policies that propose learning a flow-based model to generate trajectories in arbitrary Riemannian manifolds.
Their work showed promising results on the LASA dataset. In this work, we focus on robotic manipulation and the combination of Flow Matching with an SE(3) Invariant Transformer architecture, enabling SE(3) equivariant action generation with successful application to real robot experiments on challenging manipulation tasks.

\section{Conclusion}
\label{sec:conclusion}
We presented ActionFlow, a new policy class for robot learning from demonstrations.
On the representation level, ActionFlow consists of an SE(3) Invariant Transformer equipped with geometry-aware Invariant Point Attention.
Actions are generated using Flow Matching, a new generative model capable of obtaining high-quality samples with low inference times.
The resulting policies are fast and efficient, represent actions and observations in one common space, and yield SE(3) equivariant action generation.
Our experiments underline the effectiveness of ActionFlow's individual components and demonstrate its capabilities for solving real robotic manipulation tasks.
In the future, we would like to explore ActionFlow's capabilities for solving longer-horizon manipulation tasks.

\bibliographystyle{IEEEtran}
\bibliography{bibliography}

\begin{thebibliography}{10}
\providecommand{\url}[1]{#1}
\csname url@samestyle\endcsname
\providecommand{\newblock}{\relax}
\providecommand{\bibinfo}[2]{#2}
\providecommand{\BIBentrySTDinterwordspacing}{\spaceskip=0pt\relax}
\providecommand{\BIBentryALTinterwordstretchfactor}{4}
\providecommand{\BIBentryALTinterwordspacing}{\spaceskip=\fontdimen2\font plus
\BIBentryALTinterwordstretchfactor\fontdimen3\font minus
  \fontdimen4\font\relax}
\providecommand{\BIBforeignlanguage}[2]{{%
\expandafter\ifx\csname l@#1\endcsname\relax
\typeout{** WARNING: IEEEtran.bst: No hyphenation pattern has been}%
\typeout{** loaded for the language `#1'. Using the pattern for}%
\typeout{** the default language instead.}%
\else
\language=\csname l@#1\endcsname
\fi
#2}}
\providecommand{\BIBdecl}{\relax}
\BIBdecl

\bibitem{zhu2023viola}
Y.~Zhu, A.~Joshi, P.~Stone, and Y.~Zhu, ``Viola: Imitation learning for
  vision-based manipulation with object proposal priors,'' in \emph{Proceedings
  of The 6th Conference on Robot Learning}, 2023.

\bibitem{chi2023diffusion}
C.~Chi, S.~Feng, Y.~Du, Z.~Xu, E.~Cousineau, B.~Burchfiel, and S.~Song,
  ``Diffusion policy: Visuomotor policy learning via action diffusion,''
  \emph{arXiv preprint arXiv:2303.04137}, 2023.

\bibitem{brohan2022rt}
A.~Brohan, N.~Brown, J.~Carbajal, Y.~Chebotar, J.~Dabis, C.~Finn,
  K.~Gopalakrishnan, K.~Hausman, A.~Herzog, J.~Hsu \emph{et~al.}, ``Rt-1:
  Robotics transformer for real-world control at scale,'' \emph{arXiv preprint
  arXiv:2212.06817}, 2022.

\bibitem{zhao2023learning}
T.~Z. Zhao, V.~Kumar, S.~Levine, and C.~Finn, ``Learning fine-grained bimanual
  manipulation with low-cost hardware,'' \emph{arXiv preprint
  arXiv:2304.13705}, 2023.

\bibitem{zeng2021transporter}
A.~Zeng, P.~Florence, J.~Tompson, S.~Welker, J.~Chien, M.~Attarian,
  T.~Armstrong, I.~Krasin, D.~Duong, V.~Sindhwani \emph{et~al.}, ``Transporter
  networks: Rearranging the visual world for robotic manipulation,'' in
  \emph{Conference on Robot Learning}, 2021.

\bibitem{gervet2023act3d}
T.~Gervet, Z.~Xian, N.~Gkanatsios, and K.~Fragkiadaki, ``Act3d: Infinite
  resolution action detection transformer for robotic manipulation,''
  \emph{arXiv preprint arXiv:2306.17817}, 2023.

\bibitem{goyal2023rvt}
A.~Goyal, J.~Xu, Y.~Guo, V.~Blukis, Y.-W. Chao, and D.~Fox, ``Rvt: Robotic view
  transformer for 3d object manipulation,'' in \emph{Conference on Robot
  Learning}, 2023.

\bibitem{vosylius2024render}
V.~Vosylius, Y.~Seo, J.~Uru{\c{c}}, and S.~James, ``Render and diffuse:
  Aligning image and action spaces for diffusion-based behaviour cloning,''
  \emph{arXiv preprint arXiv:2405.18196}, 2024.

\bibitem{shridhar2023perceiver}
M.~Shridhar, L.~Manuelli, and D.~Fox, ``Perceiver-actor: A multi-task
  transformer for robotic manipulation,'' in \emph{Conference on Robot
  Learning}.\hskip 1em plus 0.5em minus 0.4em\relax PMLR, 2023, pp. 785--799.

\bibitem{james2022coarse}
S.~James, K.~Wada, T.~Laidlow, and A.~J. Davison, ``Coarse-to-fine q-attention:
  Efficient learning for visual robotic manipulation via discretisation,'' in
  \emph{Proceedings of the IEEE/CVF Conference on Computer Vision and Pattern
  Recognition}, 2022.

\bibitem{yang2023equivact}
J.~Yang, C.~Deng, J.~Wu, R.~Antonova, L.~Guibas, and J.~Bohg, ``Equivact: Sim
  (3)-equivariant visuomotor policies beyond rigid object manipulation,''
  \emph{arXiv preprint arXiv:2310.16050}, 2023.

\bibitem{ryu2023diffusion}
H.~Ryu, J.~Kim, J.~Chang, H.~S. Ahn, J.~Seo, T.~Kim, J.~Choi, and R.~Horowitz,
  ``Diffusion-edfs: Bi-equivariant denoising generative modeling on se (3) for
  visual robotic manipulation,'' \emph{arXiv preprint arXiv:2309.02685}, 2023.

\bibitem{huang2024leveraging}
H.~Huang, D.~Wang, A.~Tangri, R.~Walters, and R.~Platt, ``Leveraging symmetries
  in pick and place,'' \emph{The International Journal of Robotics Research},
  2024.

\bibitem{gao2024riemann}
C.~Gao, Z.~Xue, S.~Deng, T.~Liang, S.~Yang, L.~Shao, and H.~Xu, ``Riemann: Near
  real-time se (3)-equivariant robot manipulation without point cloud
  segmentation,'' \emph{arXiv preprint arXiv:2403.19460}, 2024.

\bibitem{simeonov2022neural}
A.~Simeonov, Y.~Du, A.~Tagliasacchi, J.~B. Tenenbaum, A.~Rodriguez, P.~Agrawal,
  and V.~Sitzmann, ``Neural descriptor fields: Se (3)-equivariant object
  representations for manipulation,'' in \emph{International Conference on
  Robotics and Automation (ICRA)}.\hskip 1em plus 0.5em minus 0.4em\relax IEEE,
  2022.

\bibitem{urain2023se}
J.~Urain, N.~Funk, J.~Peters, and G.~Chalvatzaki,
  ``S{E}(3)-{D}iffusion{F}ields: Learning smooth cost functions for joint grasp
  and motion optimization through diffusion,'' in \emph{IEEE International
  Conference on Robotics and Automation (ICRA)}.\hskip 1em plus 0.5em minus
  0.4em\relax IEEE, 2023.

\bibitem{calinon2016tutorial}
S.~Calinon, ``A tutorial on task-parameterized movement learning and
  retrieval,'' \emph{Intelligent service robotics}, vol.~9, pp. 1--29, 2016.

\bibitem{dreher2019learning}
C.~R. Dreher, M.~W{\"a}chter, and T.~Asfour, ``Learning object-action relations
  from bimanual human demonstration using graph networks,'' \emph{IEEE Robotics
  and Automation Letters}, vol.~5, no.~1, 2019.

\bibitem{jumper2021highly}
J.~Jumper, R.~Evans, A.~Pritzel, T.~Green, M.~Figurnov, O.~Ronneberger,
  K.~Tunyasuvunakool, R.~Bates, A.~{\v{Z}}{\'\i}dek, A.~Potapenko
  \emph{et~al.}, ``Highly accurate protein structure prediction with
  alphafold,'' \emph{Nature}, vol. 596, no. 7873, pp. 583--589, 2021.

\bibitem{yim2023fast}
J.~Yim, A.~Campbell, A.~Y. Foong, M.~Gastegger, J.~Jim{\'e}nez-Luna, S.~Lewis,
  V.~G. Satorras, B.~S. Veeling, R.~Barzilay, T.~Jaakkola \emph{et~al.}, ``Fast
  protein backbone generation with se (3) flow matching,'' \emph{arXiv preprint
  arXiv:2310.05297}, 2023.

\bibitem{yim2023se}
J.~Yim, B.~L. Trippe, V.~De~Bortoli, E.~Mathieu, A.~Doucet, R.~Barzilay, and
  T.~Jaakkola, ``Se (3) diffusion model with application to protein backbone
  generation,'' \emph{arXiv preprint arXiv:2302.02277}, 2023.

\bibitem{jing2024alphafold}
B.~Jing, B.~Berger, and T.~Jaakkola, ``Alphafold meets flow matching for
  generating protein ensembles,'' \emph{arXiv preprint arXiv:2402.04845}, 2024.

\bibitem{huguet2024sequence}
G.~Huguet, J.~Vuckovic, K.~Fatras, E.~Thibodeau-Laufer, P.~Lemos, R.~Islam,
  C.-H. Liu, J.~Rector-Brooks, T.~Akhound-Sadegh, M.~Bronstein \emph{et~al.},
  ``Sequence-augmented se (3)-flow matching for conditional protein backbone
  generation,'' \emph{arXiv preprint arXiv:2405.20313}, 2024.

\bibitem{esser2024scaling}
P.~Esser, S.~Kulal, A.~Blattmann, R.~Entezari, J.~M{\"u}ller, H.~Saini,
  Y.~Levi, D.~Lorenz, A.~Sauer, F.~Boesel \emph{et~al.}, ``Scaling rectified
  flow transformers for high-resolution image synthesis,'' \emph{arXiv preprint
  arXiv:2403.03206}, 2024.

\bibitem{lipman2022flow}
Y.~Lipman, R.~T. Chen, H.~Ben-Hamu, M.~Nickel, and M.~Le, ``Flow matching for
  generative modeling,'' \emph{arXiv preprint arXiv:2210.02747}, 2022.

\bibitem{chenflow}
R.~T. Chen, M.~FAIR, and Y.~Lipman, ``Flow matching on general geometries.''

\bibitem{chen2023riemannian}
R.~T. Chen and Y.~Lipman, ``Riemannian flow matching on general geometries,''
  \emph{arXiv preprint arXiv:2302.03660}, 2023.

\bibitem{liutkus2021relative}
A.~Liutkus, O.~C{\i}fka, S.-L. Wu, U.~Simsekli, Y.-H. Yang, and G.~Richard,
  ``Relative positional encoding for transformers with linear complexity,'' in
  \emph{International Conference on Machine Learning}, 2021.

\bibitem{liu2022flow}
X.~Liu, C.~Gong, and Q.~Liu, ``Flow straight and fast: Learning to generate and
  transfer data with rectified flow,'' \emph{arXiv preprint arXiv:2209.03003},
  2022.

\bibitem{chen2018neural}
R.~T. Chen, Y.~Rubanova, J.~Bettencourt, and D.~K. Duvenaud, ``Neural ordinary
  differential equations,'' \emph{Advances in neural information processing
  systems}, vol.~31, 2018.

\bibitem{wen2023foundationpose}
B.~Wen, W.~Yang, J.~Kautz, and S.~Birchfield, ``Foundationpose: Unified 6d pose
  estimation and tracking of novel objects,'' \emph{arXiv preprint
  arXiv:2312.08344}, 2023.

\bibitem{song2020score}
Y.~Song, J.~Sohl-Dickstein, D.~P. Kingma, A.~Kumar, S.~Ermon, and B.~Poole,
  ``Score-based generative modeling through stochastic differential
  equations,'' \emph{arXiv preprint arXiv:2011.13456}, 2020.

\bibitem{ho2020denoising}
J.~Ho, A.~Jain, and P.~Abbeel, ``Denoising diffusion probabilistic models,''
  \emph{Advances in neural information processing systems}, vol.~33, pp.
  6840--6851, 2020.

\bibitem{sola2018micro}
J.~Sola, J.~Deray, and D.~Atchuthan, ``A micro lie theory for state estimation
  in robotics,'' \emph{arXiv preprint arXiv:1812.01537}, 2018.

\bibitem{vaswani2017attention}
A.~Vaswani, N.~Shazeer, N.~Parmar, J.~Uszkoreit, L.~Jones, A.~N. Gomez,
  {\L}.~Kaiser, and I.~Polosukhin, ``Attention is all you need,''
  \emph{Advances in neural information processing systems}, vol.~30, 2017.

\bibitem{mandlekar2022matters}
A.~Mandlekar, D.~Xu, J.~Wong, S.~Nasiriany, C.~Wang, R.~Kulkarni, L.~Fei-Fei,
  S.~Savarese, Y.~Zhu, and R.~Mart{\'\i}n-Mart{\'\i}n, ``What matters in
  learning from offline human demonstrations for robot manipulation,'' in
  \emph{Conference on Robot Learning}, 2022.

\bibitem{song2021ddim}
J.~Song, C.~Meng, and S.~Ermon, ``Denoising diffusion implicit models,'' in
  \emph{9th International Conference on Learning Representations, {ICLR}},
  2021.

\bibitem{mandlekar2023mimicgen}
A.~Mandlekar, S.~Nasiriany, B.~Wen, I.~Akinola, Y.~Narang, L.~Fan, Y.~Zhu, and
  D.~Fox, ``Mimicgen: A data generation system for scalable robot learning
  using human demonstrations,'' \emph{arXiv preprint arXiv:2310.17596}, 2023.

\bibitem{he2016resnet}
K.~He, X.~Zhang, S.~Ren, and J.~Sun, ``Deep residual learning for image
  recognition,'' in \emph{{IEEE} Conference on Computer Vision and Pattern
  Recognition}.\hskip 1em plus 0.5em minus 0.4em\relax {IEEE} Computer Society,
  2016.

\bibitem{AmazonPresenter}
``Wireless presenter,''
  \url{https://www.amazon.de/Verbindung-USB-Empf%C3%A4nger-Laserpointer-Fernbedienung-Pr%C3%A4sentation/dp/B09DYBSJV1},
  [Accessed 02-09-2024].

\bibitem{Nbfigueroafranka_interactive_controllers}
``{F}ranka {I}nteractive {C}ontrollers,''
  \url{https://github.com/nbfigueroa/franka_interactive_controllers}, [Accessed
  02-09-2024].

\bibitem{deng2021vector}
C.~Deng, O.~Litany, Y.~Duan, A.~Poulenard, A.~Tagliasacchi, and L.~J. Guibas,
  ``Vector neurons: A general framework for so (3)-equivariant networks,'' in
  \emph{Proceedings of the IEEE/CVF International Conference on Computer
  Vision}, 2021, pp. 12\,200--12\,209.

\bibitem{ke20243d}
T.-W. Ke, N.~Gkanatsios, and K.~Fragkiadaki, ``3d diffuser actor: Policy
  diffusion with 3d scene representations,'' \emph{arXiv preprint
  arXiv:2402.10885}, 2024.

\bibitem{shen2023distilled}
W.~Shen, G.~Yang, A.~Yu, J.~Wong, L.~P. Kaelbling, and P.~Isola, ``Distilled
  feature fields enable few-shot language-guided manipulation,'' \emph{arXiv
  preprint arXiv:2308.07931}, 2023.

\bibitem{huang2022equivariant}
H.~Huang, D.~Wang, R.~Walters, and R.~Platt, ``Equivariant transporter
  network,'' in \emph{Robotics: Science and Systems}, 2022.

\bibitem{shridhar2022cliport}
M.~Shridhar, L.~Manuelli, and D.~Fox, ``Cliport: What and where pathways for
  robotic manipulation,'' in \emph{Conference on robot learning}, 2022.

\bibitem{lin2023mira}
Y.-C. Lin, P.~Florence, A.~Zeng, J.~T. Barron, Y.~Du, W.-C. Ma, A.~Simeonov,
  A.~R. Garcia, and P.~Isola, ``Mira: Mental imagery for robotic affordances,''
  in \emph{Conference on Robot Learning}, 2023.

\bibitem{shafiullah2022clip}
N.~M.~M. Shafiullah, C.~Paxton, L.~Pinto, S.~Chintala, and A.~Szlam,
  ``Clip-fields: Weakly supervised semantic fields for robotic memory,''
  \emph{arXiv preprint arXiv:2210.05663}, 2022.

\bibitem{pan2023tax}
C.~Pan, B.~Okorn, H.~Zhang, B.~Eisner, and D.~Held, ``Tax-pose: Task-specific
  cross-pose estimation for robot manipulation,'' in \emph{Conference on Robot
  Learning}, 2023.

\bibitem{ryu2023equivariant}
H.~Ryu, H.~in~Lee, J.-H. Lee, and J.~Choi, ``Equivariant descriptor fields:
  {SE}(3)-equivariant energy-based models for end-to-end visual robotic
  manipulation learning,'' in \emph{The Eleventh International Conference on
  Learning Representations}, 2023.

\bibitem{huang2024fourier}
H.~Huang, O.~Howell, X.~Zhu, D.~Wang, R.~Walters, and R.~Platt, ``Fourier
  transporter: Bi-equivariant robotic manipulation in 3d,'' \emph{arXiv
  preprint arXiv:2401.12046}, 2024.

\bibitem{ze20243d}
Y.~Ze, G.~Zhang, K.~Zhang, C.~Hu, M.~Wang, and H.~Xu, ``3d diffusion policy,''
  \emph{arXiv preprint arXiv:2403.03954}, 2024.

\bibitem{bose2023se}
A.~J. Bose, T.~Akhound-Sadegh, K.~Fatras, G.~Huguet, J.~Rector-Brooks, C.-H.
  Liu, A.~C. Nica, M.~Korablyov, M.~Bronstein, and A.~Tong, ``Se (3)-stochastic
  flow matching for protein backbone generation,'' \emph{arXiv preprint
  arXiv:2310.02391}, 2023.

\bibitem{zheng2023guided}
Q.~Zheng, M.~Le, N.~Shaul, Y.~Lipman, A.~Grover, and R.~T. Chen, ``Guided flows
  for generative modeling and decision making,'' \emph{arXiv preprint
  arXiv:2311.13443}, 2023.

\bibitem{janner2022planning}
M.~Janner, Y.~Du, J.~B. Tenenbaum, and S.~Levine, ``Planning with diffusion for
  flexible behavior synthesis,'' \emph{arXiv preprint arXiv:2205.09991}, 2022.

\bibitem{braun2024riemannian}
M.~Braun, N.~Jaquier, L.~Rozo, and T.~Asfour, ``Riemannian flow matching policy
  for robot motion learning,'' \emph{arXiv preprint arXiv:2403.10672}, 2024.

\bibitem{bose2024se3stochastic}
A.~J. Bose, T.~Akhound-Sadegh, G.~Huguet, K.~Fatras, J.~Rector-Brooks, C.-H.
  Liu, A.~C. Nica, M.~Korablyov, M.~Bronstein, and A.~Tong, ``Se(3)-stochastic
  flow matching for protein backbone generation,'' in \emph{The International
  Conference on Learning Representations (ICLR)}, 2024.

\end{thebibliography}

\newpage

\appendix

\subsection{Additional Details on Flow Matching in SE(3)}
\label{app:se3_fm}
In \Cref{sec:flow_se3}, we introduced Flow Matching in the Lie group SE(3). For completeness, we provide the pseudo-code for both training (\Cref{alg:fm_train}) and sampling (\Cref{alg:fm_sample}) with flow-based models.

\begin{algorithm}
\caption{Flow Matching Training for Action Poses}
\label{alg:fm_train}
\begin{algorithmic}[1]
\REPEAT
    \STATE $\vr_a, \vp_a, \mO \sim \gD$ \footnotesize{\texttt{\# Sample a batch of actions and observations from the dataset}}
    \STATE $\vr_0, \vp_0, \sim q_0$  \footnotesize{\texttt{\# Sample random pose from the initial distribution}}
    \STATE $t \sim \text{Uniform}(\{0, \dots, 1\})$ \footnotesize{\texttt{\# Sample timestep}}
    \STATE $\vr_t, \vp_t \leftarrow \vphi(\vr_0, \vp_0, \vr_a, \vp_a, t)$\footnotesize{\texttt{\# Compute flow given \Cref{eq:rlf_so3}}}
    \STATE $\dot{\vr}_t, \dot{\vp}_t \leftarrow \vu(\vr_t, \vp_t, \vr_a, \vp_a, t)$ \footnotesize{\texttt{\# Compute target vector given \Cref{eq:se3rlf}}}
    \STATE $\vv_p, \vv_r \leftarrow \vv_{\vtheta}(\vr_t, \vp_t, t, \mO)$ \footnotesize{\texttt{\# Predict vector given the learnable model}}
    \STATE $\gL = || \vv_p - \dot{\vp}_t ||^2 + || \vv_r - \dot{\vr}_t ||^2$ \footnotesize{\texttt{\# Compute L2 loss}}
    \STATE $\vtheta \leftarrow \nabla_{\vtheta} \gL$ \footnotesize{\texttt{\# Take gradient step}}
\UNTIL converged
\end{algorithmic}
\end{algorithm}

\begin{algorithm}
\caption{Sampling Action Poses from flow-based models}
\label{alg:fm_sample}
\begin{algorithmic}[1]
\REQUIRE An observation $\mO$
\STATE $\vr_t, \vp_t, \sim q_0$  \footnotesize{\texttt{\# Sample random pose from initial distribution}}
\FOR{$t= 0, \dots, 1$}
    \STATE $\vv_p, \vv_r \leftarrow \vv_{\vtheta}(\vr_t, \vp_t, t, \mO)$ \footnotesize{\texttt{\# Predict vector given the learned model}}
    \STATE $\vr_t, \vp_t \leftarrow \text{EulerStep} (\vr_t, \vp_t, \vv_p, \vv_r, \Delta t)$ \footnotesize{\texttt{\# Update the pose given \Cref{eq:sample_se3}}}
\ENDFOR
\RETURN $\vr_t, \vp_t$
\end{algorithmic}
\end{algorithm}

\subsection{Flow-based Policies in Euclidean Space}
\label{app:flow_pi}
In this section, we describe how to represent a Flow Matching based policy in the Euclidean space.
We provide the details as we actually evaluated the performance of Euclidean Flow Matching policies in \Cref{sec:exp_1}.
We decided to use Flow Matching in the Euclidean space in \Cref{sec:exp_1}, as it allowed for a fairer comparison with the other baselines.
We propose modeling a policy $\pi_{\vtheta}(\va|\vo)$ as a \textit{Continuous Normalizing Flow} (\gls{cnf})~\cite{chen2018neural} trained via \gls{cfm} (\Cref{eq:cfm}). Flow-based policies are \textbf{\textit{expressive}}, able to represent multimodal action distributions yet \textbf{\textit{simple}}.
Additionally, the training is stable, and the sampling method is simple and deterministic.
Similar to previous works~\cite{chi2023diffusion,zhao2023learning}, we represent the action space as a trajectory of future actions.

The problem in flow matching boils down to designing a conditioned flow that drives randomly sampled points to the dataset. In the following, we present a popular flow (Rectified Linear Flow) and showcase how it can be used to generate robot actions.
\\
\textit{\textbf{Rectified Linear Flow}}~\cite{liu2022flow, esser2024scaling, chenflow} propose representing the data point conditioned flow $\vphi_t(\va|\va_1)$ with a straight line from a noisy sample $\va_0 \sim \gN(\vzero,\mI)$ at $t=0$ to the datapoint $\va_1\in\gD$ at $t=1$
\begin{align}
    \va_t = \vphi_t(\va_0|\va_1) = t \va_1 + (1-t)\va_0
    \label{eq:rect_linear_flow}.
\end{align}
Then, sampling from the conditioned probability $\rho_t(\va|\va_1)$ can be easily done by sampling ${\va_0\sim\gN(\vzero,\mI)}$ and computing the flow at time $t$ with \Cref{eq:rect_linear_flow}. 
By differentiating \Cref{eq:rect_linear_flow}, the conditional vector field is $\vu_t(\va|\va_1) = \frac{\d }{\d t}\vphi_t(\va|\va_1) = \frac{\va - \va_1}{1-t}$.
Notice that $\vu_t$ is a constant velocity for any time $t$ defined by $\vu_t(\va|\va_1) = \va_1 - \va_0$.
An interesting property of the straight path given by the Rectified Linear Flow is that it will incur small errors with numerical solvers, an essential property if we aim to sample with very few iterations.

\textbf{Training.} Given a dataset $\gD:\{\va_n,\vo_n\}_{n=0}^N$, we train a context and time-conditioned vector field $\vv_{\vtheta}(\va, \vo, t)$ by regressing a designed vector field $\vu_t$
\begin{align}
    \gL(\vtheta) = \E_{(\va_1,\vc)\sim\gD, t, \rho_t(\va|\va_1)} \norm{\vv_{\vtheta}(\va,\vc, t) - \vu_t(\va|\va_1)}^2  ,
    \label{eq:fm_policy_train}
\end{align}
with $t\sim \gU[0,1]$ and $\rho_t$ and $\vu_t$ the probability path and vector field given by the Rectified Linear Flow.%

\textbf{Sampling.}
In our work, we propose sampling from $\pi_{\vtheta}(\va|\vc)$ by naively applying Euler discretization. 
We first sample from $\va_0\sim \gN(\vzero, \mI)$, and iteratively apply Euler discretization for $K$ steps
\begin{align}
    \va_{k+1} = \va_{k} + \vv_{\vtheta}(\va_k, \vc, k\Delta t)\Delta t,
\end{align}
with $\Delta t = 1/K$.
Notice that we can naively play with the iterations and the $\Delta t$ to find the optimal sampling for our task. In particular, in \Cref{sec:exp_1}, we observe that we can obtain highly accurate samples with very few iterations.

\subsection{Equivariant Generation with an Invariant Model}
\label{app:equiv_gen}

This section provides additional details explaining how exactly we obtain SE(3) equivariant action generation, given that the underlying transformer model is SE(3) invariant. 
\\
Given a policy $\pi(\mT_a|\mF_o, \mT_o)$ that generates action poses $\mT_a$, given the observation poses $\mT_o$, the policy is SE(3) equivariant if under a transformation $T_{\delta} \in SE(3)$ over the observations, the distribution over the actions is similarly transformed, i.e., $\pi(\mT_a|\mF_o, \mT_o) = \pi(T_{\delta} \mT_a|\mF_o, T_{\delta} \mT_o)$. 

Our proposed ActionFlow achieves equivariance by updating the action poses w.r.t. their own local frame. This results in equivariant action generation as long as the underlying model is invariant, as we will show in the following.
This property has been previously exploited in protein folding problems~\cite{jumper2021highly, yim2023se, yim2023fast}.

Let us consider the update rule represented in \Cref{eq:sample_se3}
\begin{align}
\begin{split}
    \vp_{k+1} &= \vp_k + \vr_k\vv_{\vtheta}(T_k, \mT_o, \mF_o, t)\Delta t \\ 
    \vr_{k+1} &= \vr_k \text{Exp}(\Delta t \vv_{\vtheta}(T_k, \mT_o, \mF_o, t))
    \label{eq:app_se_sample}
\end{split}
\end{align}
with $\vp$ being the translation in the world frame, $\vr$ the rotation matrix in the world frame, and the step length $\Delta t$.
Importantly, the current pose's rotation matrix $\vr_k$ is premultiplied to the predicted update vectors $\vv_{\vtheta}$. Therefore, the predicted update vector operates in the local frame. 
We aim for equivariant generation, such that if we apply a transformation $T_{\delta} = (\vp_{\delta}, \vr_{\delta}) \in SE(3)$ over the current pose $T_k = (\vp_k, \vr_k)$, i.e., 
\begin{align}
    T'_k = (\vp'_k, \vr'_k) = (\vr_{\delta}\vp_k + \vp_{\delta}, \vr_{\delta} \vr_k),
\end{align}
and observations $\mT'_o = T_{\delta} \mT_o$, the updated pose $T'_{k+1}$ is by definition similarly transformed, i.e., $T'_{k+1} = (\vp'_{k+1}, \vr'_{k+1}) = (\vr_{\delta} \vp_{k+1}+ \vp_{\delta}, \vr_{\delta} \vr_{k+1})$.

To showcase that for equivariant action generation, the model should be invariant, we start by considering the update equations for the transformed poses. They equate to
\begin{align}
\begin{split}
        \vp'_{k+1} &= \vp'_k + \vr'_k\vv_{\vtheta}(T'_k, \mT'_o, \mF_o, t)\Delta t \\ 
        \vr'_{k+1} &= \vr'_k \text{Exp}(\Delta t \vv_{\vtheta}(T'_k, \mT'_o, \mF_o, t)).
        \label{eq:new_sample}
\end{split}
\end{align}
By inserting the definitions for $(\vp'_k, \vr'_k)$ and $(\vp'_{k+1}, \vr'_{k+1})$ in \Cref{eq:new_sample}, we obtain
\begin{align}
\begin{split}
        \vr_{\delta}\vp_{k+1} + \vp_{\delta} &= \vr_{\delta}\vp_k + \vr_{\delta}\vr_k\vv_{\vtheta}(T'_k, \mT'_o, \mF_o, t)\Delta t + \vp_{\delta} \\
         \vr_{\delta}\vr_{k+1} &=  \vr_{\delta}\vr_k \text{Exp}(\Delta t \vv_{\vtheta}(T'_k, \mT'_o, \mF_o, t)).
\end{split}
\end{align}
We observe that we can cancel $\vp_{\delta}$ and $\vr_{\delta}$ on each side of the equations and obtain
\begin{align}
\begin{split}
        \vp_{k+1} &= \vp_k + \vr_k\vv_{\vtheta}(T'_k, \mT'_o, \mF_o, t)\Delta t \\
         \vr_{k+1} &= \vr_k \text{Exp}(\Delta t \vv_{\vtheta}(T'_k, \mT'_o, \mF_o, t)).
         \label{eq:newnew}
\end{split}
\end{align}
Since \Cref{eq:newnew} has to hold true for the model to generate equivariant actions, it follows (from \Cref{eq:app_se_sample}) that the model has to be invariant, i.e., $\vv_{\vtheta}(T'_k, \mT'_o, \mF_o, t) = \vv_{\vtheta}(T_k, \mT_o, \mF_o, t)$.

\subsection{Invariant Point Attention}
\label{app:ipa}

This section provides pseudo-code for the Invariant Point Attention mechanism, the key element for our SE(3) Invariant Transformer.
Since Invariant Point Attention was originally proposed in the context of protein folding \cite{jumper2021highly}, our pseudo-code in \Cref{alg:ipa_robotics} aims to provide additional context from a robotics perspective.
The algorithm receives the set of features $\mF$ and their associated poses $\mT$ as input.
The other inputs are hyper parameters.
The algorithm describes the update for one token with associated feature vector $\vf_i$ and pose $T_i$.
Implementation-wise, we built on top of~\cite{bose2024se3stochastic}.

\begin{algorithm}
\caption{Invariant Point Attention ($\mF{=}\{\vf_1, \cdots , \vf_i, \cdots \} , \mT{=}\{T_1, \cdots , T_i, \cdots \}, \\ {N_{head}}, {c}, {N_\text{query points}}, {N_\text{point values}}$)}
\label{alg:ipa_robotics}
\begin{algorithmic}[1]
    \STATE $\vq_i^h, \vk_i^h, \vv_i^h {=} \text{Linear}(\vf_i)$, $\vq_i^h, \vk_i^h, \vv_i^h {\in} \mathbb{R}^{c}, h {\in} \{1, ..., N_{head}\}$ \footnotesize{\texttt{\# Create global query, key and velocity points. c is the dimension of the embedding space.}}
    \STATE $\vec{\vq}_i^{h,p}, \vec{\vk}_i^{h,p} {=} \text{Linear}(\vf_i)$, $\vec{\vq}_i^{h,p}, \vec{\vk}_i^{h,p} {\in} \mathbb{R}^{3}, p {\in} \{1, ..., N_\text{query points}\}$ \footnotesize{\texttt{\# Create a set of three dimensional query and key points for compatibility}}
    \STATE $\vec{\vv}_i^{h,p} {=} \text{Linear}(\vf_i)$, $\vec{\vv}_i^{h,p} {\in} \mathbb{R}^{3}, p {\in} \{1, ..., N_\text{point values}\}$ \footnotesize{\texttt{\# Create a set of three dimensional feature points}}
    \STATE $w_L = \frac{1}{\sqrt{3c}}$ \quad $w_c = 0.5 \sqrt{\frac{2}{(3\cdot9)N_\text{query points}}}$ \footnotesize{\texttt{\# Initialize regularisation constants}}
    \STATE $a_{i,j}^{h} {=} \text{softmax}_j({w_L \vq_i^h}^T \vk_j^h - w_c \sum_{p}^{} \left\Vert \vr_i \vec{\vq}_i^{h,p} - \vr_j \vec{\vk}_j^{h,p} \right\Vert^2)$ \footnotesize{\texttt{\#Compute Compatibility}}
    \STATE $\vo_i^{h} {=} \sum_{j}^{} a_{i,j}^{h} \vv_j^{h}$ \footnotesize{\texttt{\#Compute how the Neighbors contribute to the update of the current Node considering the global feature which does not take into account their relative transform}}
    \STATE $\vec{\vo}_i^{h,p} {=} \vr_i^T ( \sum_{j}^{} a_{i,j}^{h} \vr_j \vec{\vv}_j^{h,p}) $ \footnotesize{\texttt{\#Compute how the Neighbors contribute to the update of the current Node considering the local ```point'' features}}
    \STATE $\tilde{\vf_i} {=} \text{Linear}(\text{concat}_{h,p}(\vo_i^{h}, \vec{\vo}_i^{h,p}, \left\Vert \vec{\vo}_i^{h,p} \right\Vert)) $ \footnotesize{\texttt{\#Compute the updated feature value for token i}}
    
\end{algorithmic}
\end{algorithm}

\subsection{Robomimic Experiments}
\label{app:robomimic_experiments}

\begin{figure}
\centering
\begin{minipage}{\columnwidth}
    \centering
    \includegraphics[width=.99\textwidth]{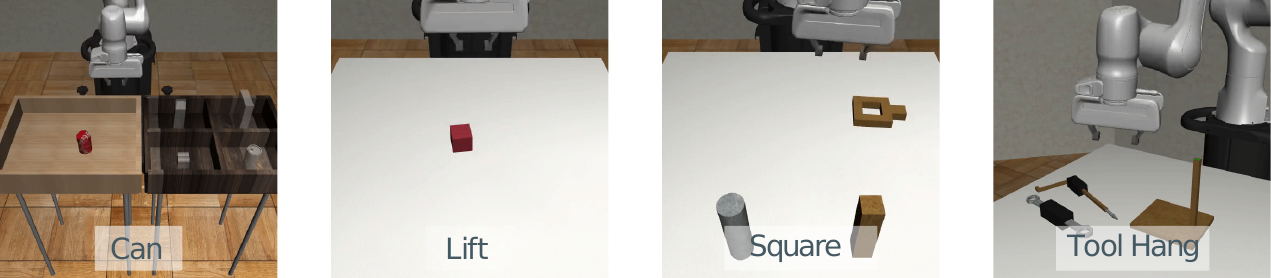}
\end{minipage}
\caption{
\textbf{Robomimic tasks} used to compare diffusion policy and flow matching in simulation.  
}
\label{fig:robomimic_environments_flat}
\end{figure}

In this section, we present more detailed results obtained in four Robomimic tasks (cf. \Cref{fig:robomimic_environments_flat})~\cite{mandlekar2022matters} using both state and image-based observations.
Robomimic contains human demonstrations of several robotic manipulation tasks in simulated environments.

As mentioned in the main text, in these experiments, the diffusion process in Diffusion Policy~\cite{chi2023diffusion} is replaced with a flow matching process in Euclidean space (\Cref{app:flow_pi}) -- we refer to this policy as \textit{Flow Matching}.
We use the transformer architecture from~\cite{chi2023diffusion} to model both the flow matching vector field and the denoising diffusion model.
We chose this architecture over the convolutional neural network since, in this work, we heavily use this type of architecture.
For a fair comparison, we use the same hyperparameters in Flow Matching as the ones provided by the Diffusion Policy authors and build our Flow Matching code on top of the authors' code from~\url{https://github.com/real-stanford/diffusion_policy}.

\textbf{Training.}
Diffusion Policy is trained using DDPM~\cite{ho2020denoising} and cosine schedule with $K=100$ steps.
Flow Matching uses the rectified linear flow (\Cref{eq:rect_linear_flow}) and also $K=100$ steps.
Due to limited computing resources, both policies are trained to a maximum of $3$ days or $4000$ epochs for the dataset with state-based observations.
For the dataset with image-based observations, both policies are trained to a maximum of $3$ days or $3500$ epochs.
Checkpoints are evaluated every $100$ epochs.
We use a machine with the following components (CPU, RAM, GPU): AMD EPYC 7453 28-Core; 512 GB RAM; RTX 3090 Turbo (24 GB).

\textbf{Inference.}
For testing the policies, we choose the (epoch) checkpoint that performs best during training, i.e., with the highest average success rate.
We report the average success rate from policy rollouts from $50$ different initial configurations (from the test set) across $3$ seeds.
During inference, we use fewer steps than during training since it allows policies to be run at higher frequencies.
For Diffusion Policy, we use DDIM for faster sampling~\cite{song2021ddim}.
For Flow Matching, we use exponentially spaced time steps (from the linear training schedule).
This has the benefit of using larger Euler integration steps $\Delta t$ when $t$ is closer to $0$ and smaller ones when it is closer to $1$.
We report the success rates obtained by both methods when using the same number of calls to the vector field or the denoising model (number of inference steps in \Cref{table:experiment_robomimic_low_dim_table,table:experiment_robomimic_image_table}).

\textbf{Results.}
\Cref{table:experiment_robomimic_low_dim_table,table:experiment_robomimic_image_table} underline the findings from the main text: flow matching-based policies are capable of obtaining comparable results with diffusion-based policies with both state and image-based observations.
In the tables, we highlight the rows in bold, which correspond to using a very low number of inference steps (i.e., $2$ steps).
Low inference steps allow for higher policy frequencies, e.g., $2$ steps correspond to approximately $100$~Hz.
We observe that in this low inference steps regime, the success rate from flow matching policies surpasses that from diffusion policies, particularly in tasks such as Square and Tool Hang, where precise actions are needed to complete the tasks.

\begin{table*}
\centering
\begin{minipage}{.99\textwidth}
    \resizebox{.99\textwidth}{!}{%
    \centering
    \begin{tblr}{
        colspec={l c | cc  cc  cc  c },
        font=\footnotesize,
        row{1-2}={font=\bfseries},
        row{3, 8}={font=\boldmath\bfseries},
        row{3-7} = {bg=gray!10},
        hline{2} = {3-20}{solid},
        width=\linewidth,
    }
        \SetCell[c=1,r=2]{l} \textbf{Method} & \SetCell[c=1,r=2]{l} \textbf{IS} & \SetCell[c=2]{c} Can & & \SetCell[c=2]{c} Lift  & &  \SetCell[c=2]{c} Square & & \SetCell[c=1]{c} ToolHang  \\
        & &   ph & mh  &  ph & mh  &    ph & mh  &   ph \\
        \hline
        \SetCell[c=1,r=5]{l} Diffusion Policy
            & 2 & $0.93 \pm 0.03$ & $0.94 \pm 0.03$ & $1.00 \pm 0.00$ & $1.00 \pm 0.00$ & $0.72 \pm 0.11$ & $0.65 \pm 0.01$ & $0.51 \pm 0.15$ \\
            & 5 & $0.99 \pm 0.02$ & $0.96 \pm 0.03$ & $1.00 \pm 0.00$ & $1.00 \pm 0.00$ & $0.90 \pm 0.03$ & $0.77 \pm 0.03$ & $0.81 \pm 0.08$ \\
            & 10 & $0.98 \pm 0.03$ & $0.97 \pm 0.01$ & $1.00 \pm 0.00$ & $1.00 \pm 0.00$ & $0.91 \pm 0.02$ & $0.79 \pm 0.06$ & $0.90 \pm 0.02$ \\
            & 20 & $0.99 \pm 0.02$ & $0.96 \pm 0.02$ & $1.00 \pm 0.00$ & $1.00 \pm 0.00$ & $0.86 \pm 0.06$ & $0.74 \pm 0.02$ & $0.90 \pm 0.03$ \\
            & 100 & $0.98 \pm 0.03$ & $0.95 \pm 0.03$ & $1.00 \pm 0.00$ & $1.00 \pm 0.00$ & $0.91 \pm 0.02$ & $0.67 \pm 0.10$ & $0.85 \pm 0.05$ \\
         \hline
        \SetCell[c=1,r=5]{l} \textbf{Flow Matching} 
            & 2 & $0.99 \pm 0.01$ & $0.97 \pm 0.02$ & $1.00 \pm 0.00$ & $1.00 \pm 0.00$ & $0.83 \pm 0.05$ & $0.73 \pm 0.06$ & $0.73 \pm 0.11$ \\
            & 5 & $0.98 \pm 0.00$ & $0.95 \pm 0.06$ & $1.00 \pm 0.00$ & $1.00 \pm 0.00$ & $0.91 \pm 0.07$ & $0.73 \pm 0.08$ & $0.85 \pm 0.03$ \\
            & 10 & $0.99 \pm 0.01$ & $0.96 \pm 0.02$ & $0.99 \pm 0.01$ & $1.00 \pm 0.00$ & $0.87 \pm 0.10$ & $0.65 \pm 0.03$ & $0.81 \pm 0.09$ \\
            & 20 & $0.99 \pm 0.01$ & $0.96 \pm 0.03$ & $0.99 \pm 0.01$ & $1.00 \pm 0.00$ & $0.85 \pm 0.06$ & $0.65 \pm 0.07$ & $0.85 \pm 0.05$ \\
            & 100 & $0.99 \pm 0.01$ & $0.96 \pm 0.03$ & $0.99 \pm 0.01$ & $1.00 \pm 0.00$ & $0.88 \pm 0.05$ & $0.67 \pm 0.06$ & $0.87 \pm 0.03$ \\
        \hline
    \end{tblr}
    }
\end{minipage}
\caption{
\textbf{Robomimic results for policies using state-based observations.}
Success rate (mean $\pm$ std) evaluation on Robomimic tasks with proprioception observations averaged over 3 seeds and 50 environments initializations (in the test set).
The models are trained with $100$ steps and tested with different inference steps (IS).
Diffusion Policy models are trained with DDPM, and inference is done with DDIM, except for IS=$100$, for which we use DDPM since these are the number of steps the model was trained on.
}
\label{table:experiment_robomimic_low_dim_table}
\end{table*}

\begin{table*}
\centering
\begin{minipage}{.99\textwidth}
    \resizebox{.99\textwidth}{!}{%
    \centering
    \begin{tblr}{
        colspec={l c | cc  cc  cc  c },
        font=\footnotesize,
        row{1-2}={font=\bfseries},
        row{3, 8}={font=\boldmath\bfseries},
        row{3-7} = {bg=gray!10},
        hline{2} = {3-20}{solid},
        width=\linewidth,
    }
        \SetCell[c=1,r=2]{l} \textbf{Method} & \SetCell[c=1,r=2]{l} \textbf{IS} & \SetCell[c=2]{c} Can & & \SetCell[c=2]{c} Lift  & &  \SetCell[c=2]{c} Square & & \SetCell[c=1]{c} ToolHang  \\
        & &   ph & mh  &  ph & mh  &    ph & mh  &   ph \\
        \hline
        \SetCell[c=1,r=5]{l} Diffusion Policy
             & 2 & $0.93 \pm 0.06$ & $0.88 \pm 0.04$ & $1.00 \pm 0.00$ & $1.00 \pm 0.00$ & $0.85 \pm 0.03$ & $0.67 \pm 0.10$ & $0.15 \pm 0.07$ \\
             & 5 & $0.95 \pm 0.05$ & $0.93 \pm 0.01$ & $1.00 \pm 0.00$ & $0.99 \pm 0.01$ & $0.89 \pm 0.02$ & $0.73 \pm 0.10$ & $0.54 \pm 0.14$ \\
             & 10 & $0.95 \pm 0.04$ & $0.91 \pm 0.01$ & $1.00 \pm 0.00$ & $1.00 \pm 0.00$ & $0.88 \pm 0.03$ & $0.81 \pm 0.08$ & $0.64 \pm 0.12$ \\
             & 20 & $0.96 \pm 0.03$ & $0.92 \pm 0.02$ & $1.00 \pm 0.00$ & $0.99 \pm 0.01$ & $0.87 \pm 0.04$ & $0.76 \pm 0.07$ & $0.68 \pm 0.13$ \\
             & 100 & $0.97 \pm 0.04$ & $0.92 \pm 0.03$ & $1.00 \pm 0.00$ & $0.99 \pm 0.01$ & $0.90 \pm 0.04$ & $0.75 \pm 0.07$ & $0.64 \pm 0.09$ \\
         \hline
        \SetCell[c=1,r=5]{l} \textbf{Flow Matching} 
             & 2 & $0.95 \pm 0.01$ & $0.95 \pm 0.01$ & $1.00 \pm 0.00$ & $1.00 \pm 0.00$ & $0.93 \pm 0.03$ & $0.73 \pm 0.01$ & $0.43 \pm 0.05$ \\
             & 5 & $0.96 \pm 0.02$ & $0.93 \pm 0.02$ & $1.00 \pm 0.00$ & $1.00 \pm 0.00$ & $0.94 \pm 0.04$ & $0.71 \pm 0.08$ & $0.47 \pm 0.03$ \\
             & 10 & $0.96 \pm 0.02$ & $0.95 \pm 0.01$ & $1.00 \pm 0.00$ & $1.00 \pm 0.00$ & $0.93 \pm 0.02$ & $0.72 \pm 0.14$ & $0.51 \pm 0.11$ \\
             & 20 & $0.95 \pm 0.03$ & $0.91 \pm 0.04$ & $1.00 \pm 0.00$ & $1.00 \pm 0.00$ & $0.95 \pm 0.02$ & $0.73 \pm 0.13$ & $0.57 \pm 0.13$ \\
             & 100 & $0.97 \pm 0.01$ & $0.93 \pm 0.03$ & $1.00 \pm 0.00$ & $1.00 \pm 0.00$ & $0.92 \pm 0.05$ & $0.74 \pm 0.09$ & $0.58 \pm 0.12$ \\
        \hline
    \end{tblr}
    }
\end{minipage}
\caption{
\textbf{Robomimic results for policies using image-based observations.}
Success rate (mean $\pm$ std) evaluation on Robomimic tasks with image observations averaged over 3 seeds and 50 environments initializations (in the test set).
The models are trained with $100$ steps and tested with different inference steps (IS).
Diffusion Policy models are trained with DDPM, and inference is done with DDIM, except for IS=$100$, for which we use DDPM since these are the number of steps the model was trained on.
}
\label{table:experiment_robomimic_image_table}
\end{table*}

\begin{figure*}
\begin{minipage}{.24\textwidth}
    \resizebox{.99\textwidth}{!}{%
    \centering
    \includegraphics[width=.99\textwidth]{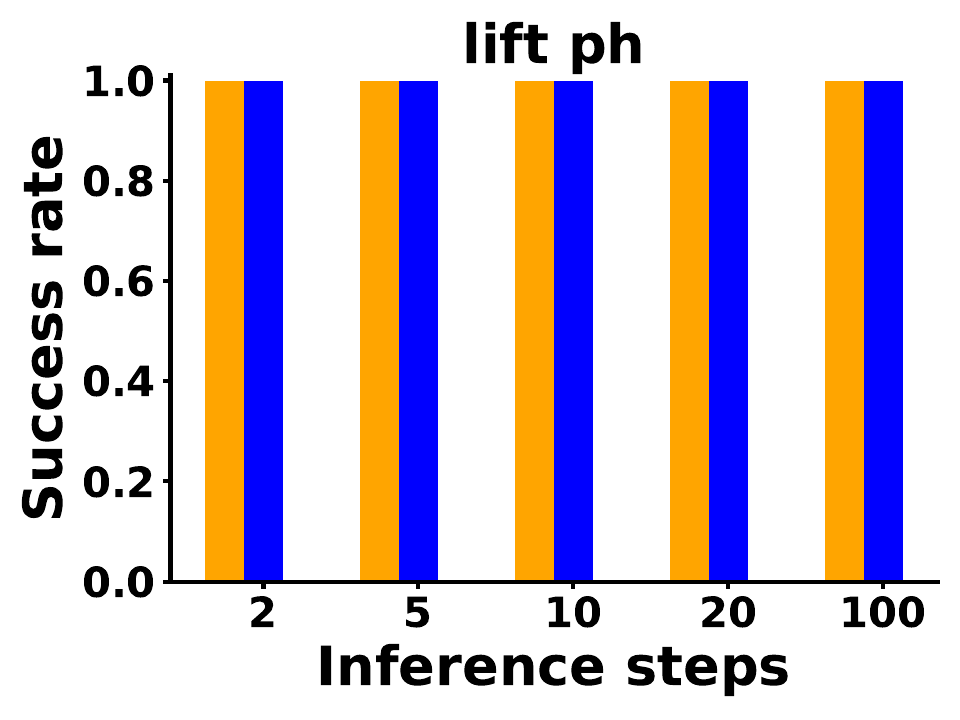}
    }
\end{minipage}
\begin{minipage}{.24\textwidth}
    \resizebox{.99\textwidth}{!}{%
    \centering
    \includegraphics[width=.99\textwidth]{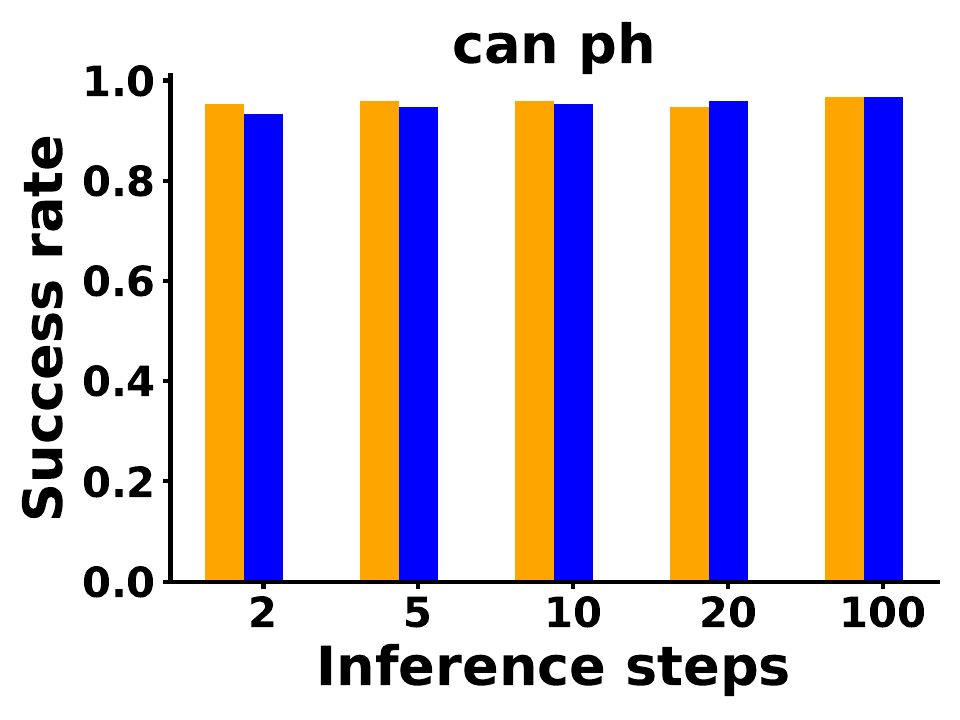}
    }
\end{minipage}
\begin{minipage}{.24\textwidth}
    \resizebox{.99\textwidth}{!}{%
    \centering
    \includegraphics[width=.99\textwidth]{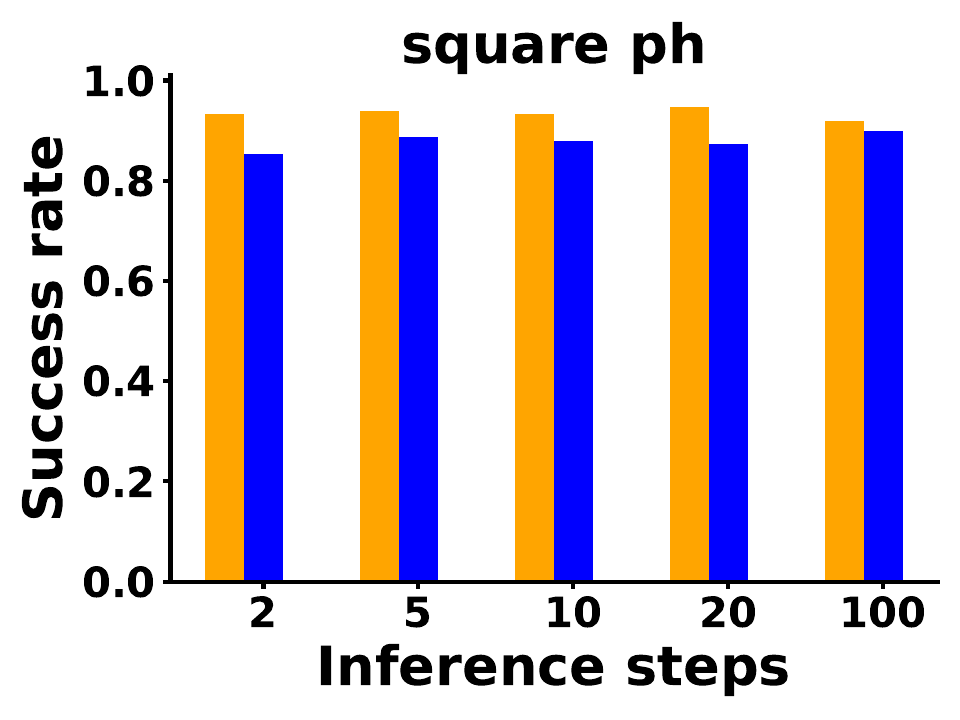}
    }
\end{minipage}
\\
\begin{minipage}{.24\textwidth}
    \resizebox{.99\textwidth}{!}{%
    \centering
    \includegraphics[width=.99\textwidth]{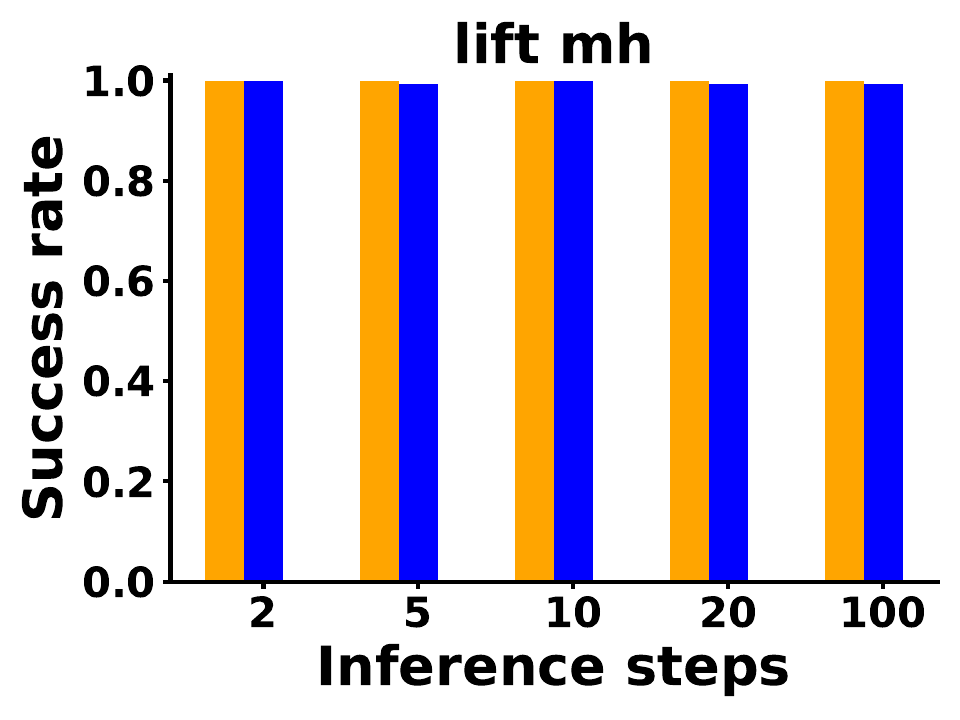}
    }
\end{minipage}
\begin{minipage}{.24\textwidth}
    \resizebox{.99\textwidth}{!}{%
    \centering
    \includegraphics[width=.99\textwidth]{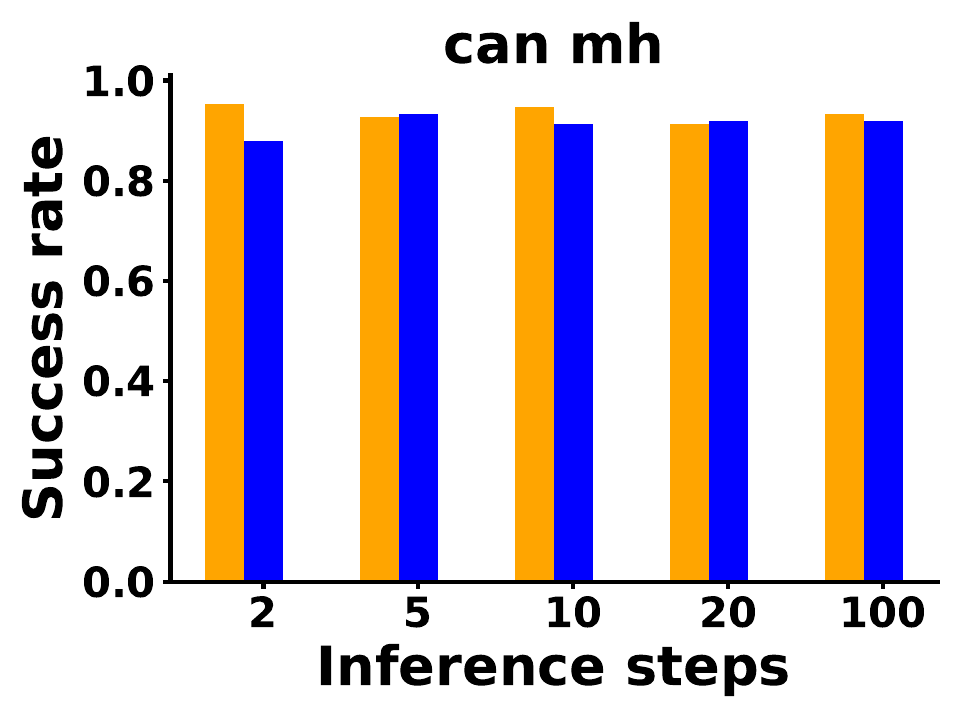}
    }
\end{minipage}
\begin{minipage}{.24\textwidth}
    \resizebox{.99\textwidth}{!}{%
    \centering
    \includegraphics[width=.99\textwidth]{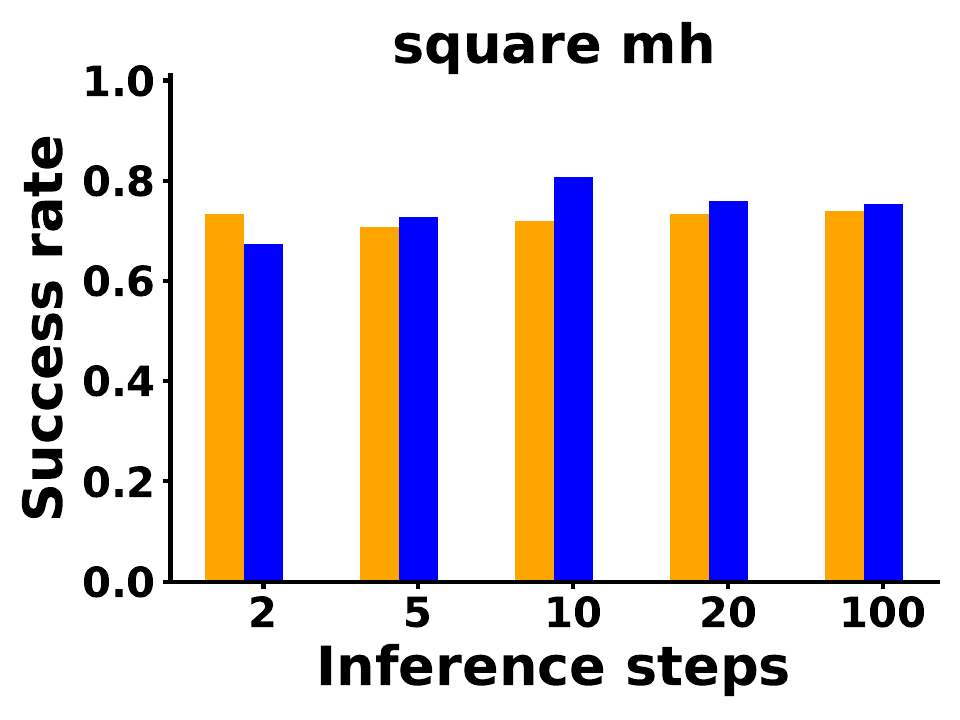}
    }
\end{minipage}
\begin{minipage}{.24\textwidth}
    \resizebox{.99\textwidth}{!}{%
    \centering
    \includegraphics[width=.99\textwidth]{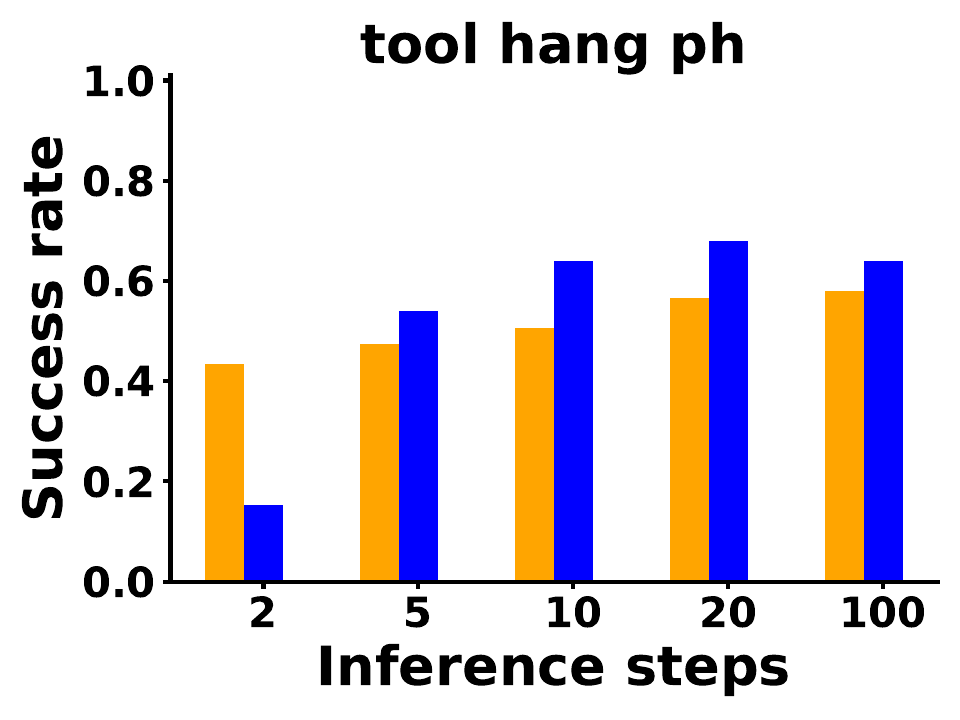}
    }
\end{minipage}
\\
\begin{minipage}{0.99\textwidth}
    \centering
    \resizebox{.35\textwidth}{!}{%
    \includegraphics[width=1.0\textwidth]{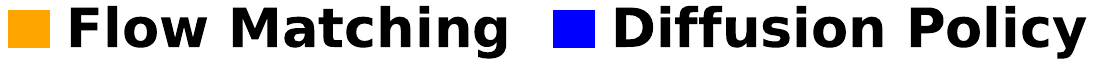}
    }
\end{minipage}
\caption{
\textbf{Robomimic results for policies using image-based observations}.
The bar plots showcase the mean results from~\Cref{table:experiment_robomimic_image_table} and provide a better visualization.
}
\label{fig:experiment_robomimic_image_plot}
\end{figure*}

\subsection{Mimicgen Experiments}
\label{app:obs_act_rep}

In this section, we provide a more detailed description of the experiment introduced in \Cref{sec:exp_2}.

\subsubsection{Observations and Actions Representation}
ActionFlow represents both the observations and actions with a tuple of poses $\mT$ and features $\mF$.
Each pair of pose and feature represents a different entity in the space.
For the experiments in \Cref{sec:exp_2}, the pose represents the location of the different relevant objects in the space, while the feature is a fixed identifier for each object. The considered objects in each task are:
\\
\textbf{Three Piece Assembly:} robot's end effector, the base, piece 1, and piece 2.
\\
\textbf{Coffee:} robot's end effector, coffee pod, coffee machine, coffee pod holder, and coffee machine lid.
\\
\textbf{Stack Three:} robot's end effector, cubeA, cubeB, and cubeC.
\\
\textbf{Threading:} robot's end effector, needle, and tripod.
\\
Additionally, the action is also expressed as a pose, representing the desired target pose for the robot's end-effector.

\subsubsection{Policy Representation}

ActionFlow's network is a SE(3) Invariant Transformer as introduced in \Cref{sec:se3_inv_trans}.
We additionally introduce an adaptation and normalization module, which is applied to the poses before they are further processed inside the transformer network. This is a common practice when training deep learning models.
\\
\textbf{Adaptation Module.} 
Given a set of observation and action poses, we represent all the poses in the end-effector's frame. 
Then, we scale the translation vectors with a scaling factor of $10$ and apply $Tanh$ to the translations to regularize the distances to a range within $-1$ and $1$.
The objective of this adaptation module is to increase the distance error between the points in the \gls{ipa} module, while reducing variability when the object's are too far. 
We represent all the poses around the end effector to guarantee that the initial distribution of the flow is centered close to the actions.

\subsection{Real Robot Experiments}
\label{app:reaL_robot}

This section provides additional details and results regarding our real robot experiments presented in \Cref{sec:rr_exps}.

\subsubsection{Additional Information regarding the Teleoperation Interface}

The teleoperation interface used in our real robot experiments consists of two main components.
We leverage an off-the-shelf presenter \cite{AmazonPresenter} for conveniently starting and stopping the recording of the individual demonstrations, as well as controlling, i.e., opening and closing the gripper.
To control the pose of the robot's end effector, we rely on the OptiTrack Motion capture system.
In particular, as shown in \Cref{fig:teleop_lightbulb_mounting}, the teleoperator wears a glove that has OptiTrack markers rigidly attached to it.
Upon starting teleoperation, the glove's current pose is defined as the reference.
Moving the glove w.r.t.~this reference results in moving the robot's end-effector w.r.t.~its initial pose accordingly.
The teleoperation interface is set to operate at \SI{25}{\hertz}.
Throughout all the real robot experiments, we use this teleoperation interface to control the robot end-effector's 6D pose, as well as the gripper opening width through a binary signal corresponding to gripper open / closed.
Last, we want to point out that the last phase of the lightbulb mounting task solely necessitates a rotation to fix the bulb and turn it on.
We found it extremely challenging to command a pure rotation around the end-effector's upward-pointing axis through the teleoperation interface.
We, therefore, assigned one of the presenter's keys to trigger a rotation of $67^{\circ}$ around the end-effector's upward-pointing axis.
Thus, for the lightbulb mounting task, the teleoperator is mainly tasked with inserting the lightbulb's pins into the socket, and subsequently, the necessary rotation can be achieved by pressing the presenter's respective key.

\subsubsection{Robot Control}

As shown, e.g., in \Cref{fig:main} \& \Cref{fig:teleop_lightbulb_mounting}, this work uses a Franka Panda 7 DoF manipulator equipped with a parallel gripper.
On the lowest level, we control the robot through the effort joint interface.
This interface requires real-time control actions at \SI{1000}{\hertz}.
For converting the current desired end-effector pose (which is either provided through the teleoperation interface or the running policy) into the effort joint commands, we build on top of the Cartesian Pose Impedance Controller provided in~\cite{Nbfigueroafranka_interactive_controllers}.
Since we do not have any smoothness guarantees on the output of our teleoperation interface and the policy's output, we employ exponential smoothing on the update of the desired target pose.
In practice, we found this measure sufficient to stay within the Franka Panda robot's acceleration limits and yield smooth trajectories for both teleoperation and policy rollouts.

\subsubsection{ActionFlow for Real Robot Manipulation - Implementation Details}

The real robot experiments presented in \Cref{sec:rr_exps} are conducted using ActionFlow, i.e., the combination of the proposed SE(3) Invariant Transformer and SE(3) Flow Matching on the action space resulting in equivariant action generation.

\textbf{Observations \& Actions.} For both experiments, we use an observation history of 5 steps and predict an action sequence containing 16 steps.
In line with the teleoperation interface (which is set to collect actions at \SI{25}{\hertz}), we employ a time discretization of \SI{0.04}{\s}.
While the RealSense camera returns RGB readings with a resolution of $640 \times 480$, we resize the images to $80 \times 80$ pixels before passing them through the ResNet18 \cite{he2016resnet} for obtaining the encodings.
The resizing helps to reduce the dataset's size significantly and, therefore, facilitates \& speeds up policy training.

\textbf{Training.}
We parametrize our ActionFlow policies for real robot manipulation using the SE(3) Invariant Transformer introduced in \Cref{fig:se3_invariant_transformer}, and use four layers of \gls{ipa}.
Additionally, we consider $K=4$ inference steps.
We train the policies for $75$ epochs and evaluate the last checkpoint.
We use a machine with the following components (CPU, RAM, GPU): AMD EPYC 7453 28-Core; 512 GB RAM; RTX 3090 Turbo (24 GB).

\textbf{Inference.}
As mentioned in \Cref{sec:rr_exps}, our policies are efficient and can be run in real-time.
On average, it takes \SI{0.03}{\s} to generate an action sequence of 16 steps on an NVIDIA RTX $3090$ GPU.
On the real robot, we also account for the delay between passing the observations to the model and obtaining the action sequences.
This is done by monitoring the time required for model inference and skipping the respective entries within the action sequence.
In particular, we skip the actions that should have been applied at times when the model inference was still active.
Moreover, we do not apply the whole remaining action sequence after each call to the model. 
Instead, we leverage our model's fast inference speeds and only apply the two next actions.
Additionally, we employ exponential smoothing to ensure a smooth transition when updating the action sequence.

\textbf{Modified Image Observations for the Mug Hanging Task.}
The mug hanging task can be divided into two phases, i.e., 1) reaching and grasping the mug, and subsequently, 2) hanging it.
While for the first part, the view of the robot's wrist-mounted camera is essential, for the second phase of approaching the hanger, apart from the mug's pose, the camera image does not contain any information.
In initial experiments, we nevertheless found that the ResNet18 extracts spurious correlations from background pixels for the phase of approaching the hanger, which harmed performance.
To circumvent this issue, once the robot's gripper is closed, we set all pixels to black apart from the image's center region of size $30 \times 30$ pixels.
This additional inductive bias, i.e., eliminating all the image background information once the mug is grasped, effectively improved the models' performance.
We want to point out that a similar effect could be achieved by masking out background pixels based on the camera's depth information. However, we discovered that the depth readings from the Intel RealSense D435 were not accurate enough for this purpose, so we decided to employ the previously described image masking once the gripper is closed.

\subsubsection{Additional Information on the Point Cloud conditioned Mug Hanging Experiment}

\begin{figure*}[t]
\centering
\includegraphics[width=.99\textwidth]{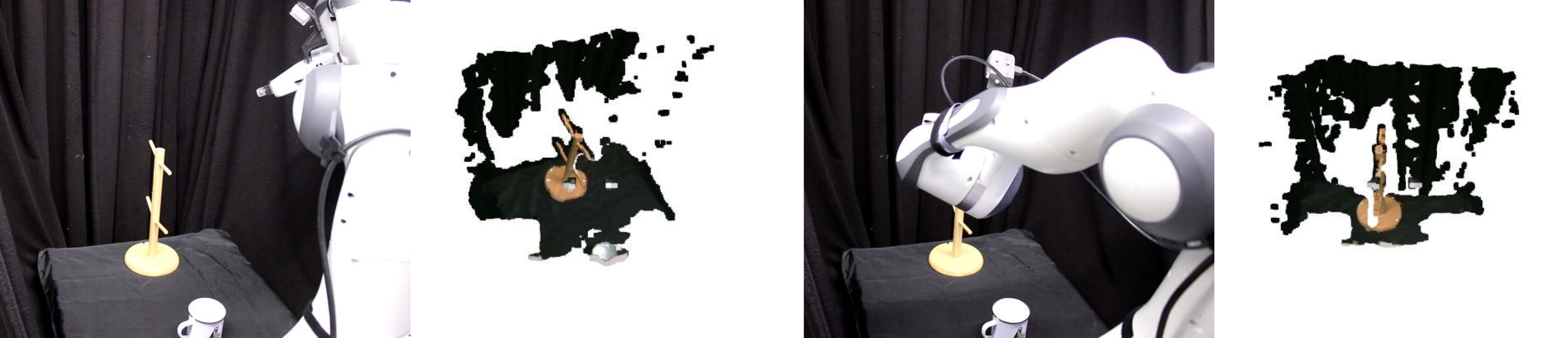}
\caption{
\textbf{}Illustrating the initial phase for the point cloud-conditioned mug hanging experiment in which a couple of waypoints are visited to register the hanger. Shown are two of these waypoints and the corresponding raw pointcloud observations next to them.}
\label{fig:pcl_initial_hovering}
\end{figure*}

This last section provides additional details about the mug hanging experiment presented in \Cref{sec:rr_exps}.
For the point cloud conditioned mug hanging experiment, we had to switch from the RealSense D435 which is used in all other experiments to an RealSense D405.
The reason for this change in RGB-D camera is that initial testing revealed that the depth readings from the RealSense D435 are insufficient to capture the details of the thin hanger.
Therefore, the point cloud conditioned version of this experiment used a RealSense D405, which is mounted in the same pose as the RealSense D435 has been mounted in the previous experiments.
Moreover, as mentioned in the main paper, to obtain a good point cloud of the hanger, before starting the teleoperation (for data collection) or the policy rollout, the robot visits 7 pre-defined end-effector poses, which ensure good visibility of the hanger as shown in \Cref{fig:pcl_initial_hovering}.

\subsubsection{Additional Information on the Pose-based Mug Hanging Experiment}

\begin{figure}[t]
\begin{center}
\begin{minipage}{0.65\columnwidth}
    \resizebox{.99\textwidth}{!}{%
    \centering
    \begin{tblr}{
        colspec={l | c },
        row{1}={font=\bfseries\footnotesize},
        row{2-20}={font=\footnotesize},
        row{3-3} = {bg=gray!10},
        width=\textwidth,
    }
        \textbf{Initialization} & \textbf{ActionFlow Success Rate} \\
        \hline
        Train & 9/10 \\
        Test &  8/10 \\
    \end{tblr}
    }
\end{minipage}
\end{center}
\vspace{-0.3cm}
\caption{Success Rates for the Pose-conditioned Mug hanging experiment on train and test configurations.}
\label{fig:app_experiment_mug_hang}
\vspace{-0.45cm}
\end{figure}

\begin{figure*}[t]
\centering
    \includegraphics[width=.99\textwidth]{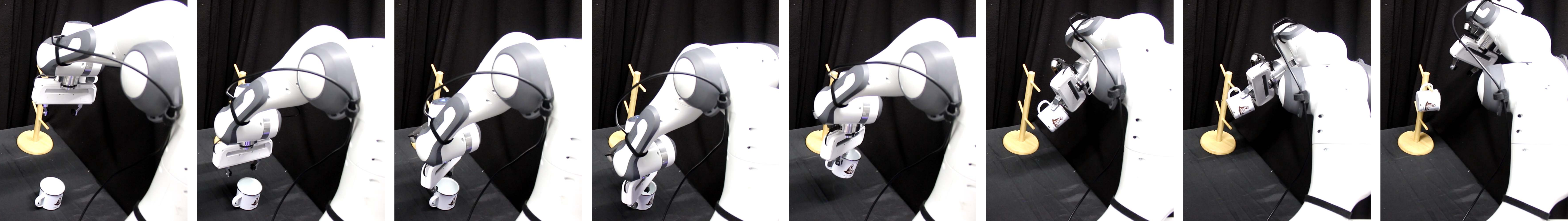}
\caption{Visualization of one ActionFlow policy rollout on the mug hanging task, in which the hanger is placed in a pose that was not included in the demonstrations.}
\label{fig:app_experiment_mug_hang_rollout}
\vspace{-0.5cm}
\end{figure*}

As mentioned in the main paper, for the task of picking up a mug and placing it onto a hanger, we also trained an ActionFlow policy that directly obtains the hanger's pose from OptiTrack readings.
Therefore, this baseline policy is conditioned on the RGB images from the wrist camera and the hanger's pose obtained through OptiTrack.
For training this baseline policy, we collect 200 demonstrations using variations, as shown in the main paper.
To reiterate, the demonstrations only include slight variations of the mug poses, while the hanger always stays in the same pose.
The results are presented in \Cref{fig:app_experiment_mug_hang}.
The table's first row reveals that the ActionFlow policy achieves high success rates of 90\% upon evaluating in similar scenarios as those encountered during training.
We only observe one failure in which the mug is not grasped properly.
Importantly, our policies run online in real-time as action generation only takes \SI{0.03}{\s} on an NVIDIA RTX $3090$ GPU.
We also evaluate the policy in previously unseen test scenarios, where, as mentioned in the main paper, the hanger is moved to either side of the table.
The results show that our policy can still handle these novel test scenarios well, achieving 80\% success.
\Cref{fig:app_experiment_mug_hang_rollout} shows a policy rollout in one of the testing scenarios.
These high success rates, despite the previously unseen scenarios, demonstrate the equivariance property of our proposed ActionFlow, which inherently can handle these translated scene instances.

One potential explanation for the slightly increased success rates in the test configurations compared to the experiments presented in \Cref{sec:rr_exps} might be that despite the changed hanger pose in the test configurations, the waypoints visited during initialization do not change. Therefore, there might be a slight distribution shift w.r.t. the hanger's point cloud and its latent encoding from the point cloud encoder, which might negatively affect the network's generalization.
This could be tackled, e.g., by adding more initial waypoints to ensure better coverage of the overall space or further refining the point cloud encoder's architecture.

\end{document}